\documentclass{article}
\usepackage{times}
\usepackage{graphicx}
\usepackage{algorithm}
\usepackage{algpseudocode}
\usepackage{booktabs}
\usepackage{subcaption}
\usepackage{pgffor}
\usepackage{amsmath,amsfonts,bm}
\usepackage[final]{corl_2026} % Uncomment for the camera-ready ``final'' version.
\makeatletter\renewcommand{\@conferencelocation}{Austin, Texas, USA}\makeatother

\title{Coupled Local and Global World Models for Efficient First Order RL}
\hypersetup{pdftitle={Coupled Local and Global World Models for Efficient First Order RL},
  pdfauthor={Joseph Amigo, Rooholla Khorrambakht, Nicolas Mansard, Ludovic Righetti}}

\author{
  Joseph Amigo$^{1,2*}$, Rooholla Khorrambakht$^{1*}$,
  Nicolas Mansard$^{2,3}$, Ludovic Righetti$^{1,3}$
  \\ $^{1}$Machines in Motion Laboratory, New York University, USA
  \\ $^{2}$LAAS-CNRS, Universit\'e de Toulouse, CNRS, Toulouse, France
  \\ $^{3}$Artificial and Natural Intelligence Toulouse Institute, Toulouse, France
  \\ \url{https://coupled-global-local-wm-rl.pages.dev/}
}

\begin{document}
\maketitle

%===============================================================================
\vspace{-1cm}
\begin{figure}[h!]
    \centering
    \includegraphics[width=1.0\linewidth]{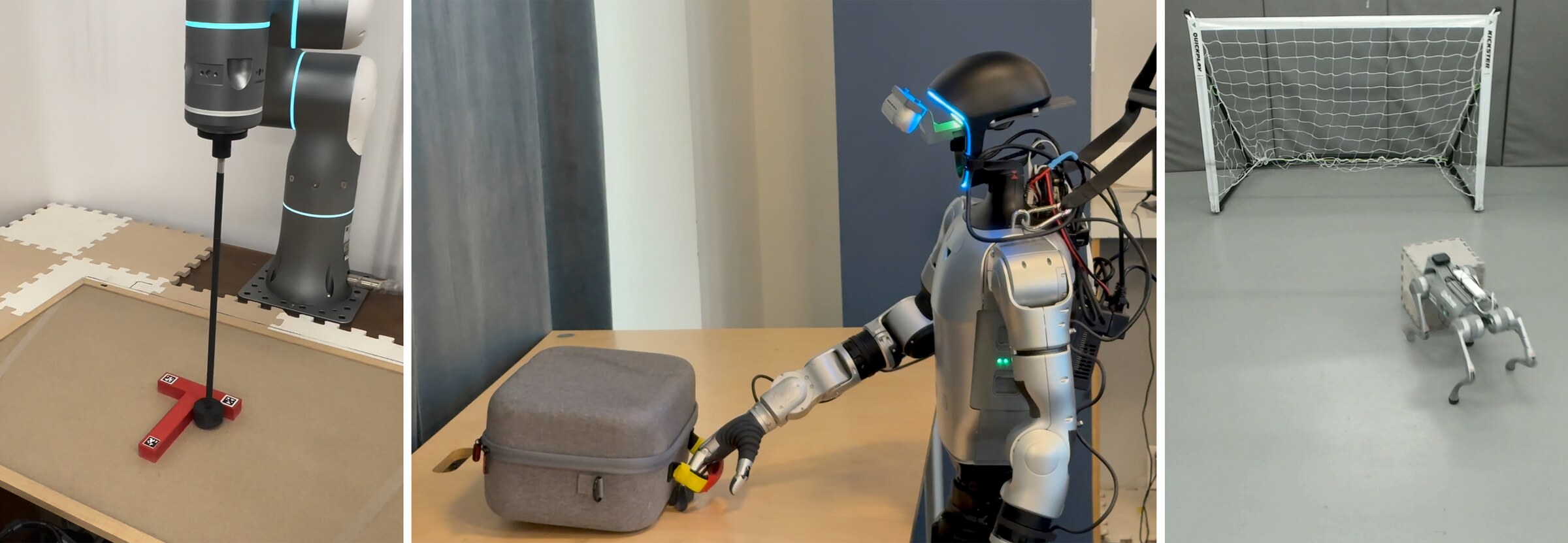}
    \caption{Real-world robotic tasks solved zero-shot by policies trained entirely from scratch within our learned world models: Push-T with a tabletop manipulator (left), Ego-Centric Grasp and Lift with a G1 humanoid (center), and Ego-Centric Push Cube with a Go2 quadruped (right).} 
    \label{fig:banner}
\end{figure}
\vspace{-0.3cm}
\begin{abstract}

    World models offer a promising avenue for capturing complex environment dynamics where simulators face challenges. However, large-scale world models required for complex real-world settings are computationally expensive to adopt in popular RL approaches. In this paper, we introduce a novel first-order RL method that enables policy training via a novel decoupled first-order gradient (FoG) method: a large-scale world model generates accurate forward trajectories, while a lightweight latent-space surrogate approximates its local dynamics for efficient gradient computation. Our coupled local and global world model formulation allows high-fidelity forward dynamics alongside computationally efficient differentiation needed for model-based RL. Through real-world experiments across a range of robotic tasks with increasing complexity, we demonstrate tractable RL and zero-shot deployment thanks to our world-model-based formulation. We show significantly better sample efficiency compared to PPO in a canonical real-world Push-T benchmark and observe similar patterns in more complicated ego-centric object manipulation and grasping tasks. Together, these results demonstrate that learning agents entirely inside data-driven world models is a feasible and promising pathway for solving hard-to-model tasks in image space without reliance on hand-crafted physics simulators.
\end{abstract}

% Two or three meaningful keywords should be added here

\renewcommand{\thefootnote}{\fnsymbol{footnote}}
\footnotetext[1]{Equal contribution.}
\renewcommand{\thefootnote}{\arabic{footnote}}

\section{Introduction}
Recent breakthroughs in reinforcement learning (RL) for locomotion \citep{cheng2023parkour, hoeller2024anymal, zhuang2023robot, chane2024cat, chane2024soloparkour} have yielded highly agile policies, trained in parallelized GPU simulators and seamlessly deployed on real robots for acrobatic tasks. Despite these gains, RL confronts substantial barriers in perceptual control and manipulation tasks. On the one hand, RL's sample inefficiency prevents practical online learning directly from real-robot interactions. On the other hand, simulating manipulation tasks is inherently more difficult than locomotion, as the environment state changes with the agent's action, adding an infinite degree of variability to the simulation problem, far beyond the ego-body dynamics and actuation modeling needed in locomotion. For instance, simulating the granular dynamics of gravel bags, pressurized deformations of an aluminum can while pulling its pull-tab, or the intricacies of folding a T-shirt, all projected onto the robot's raw sensory space, even if computationally feasible in some cases, fall short in terms of alignment with reality and sim2real transferability.

%Instead, training agents inside learned differentiable world models using first-order gradient (FoG) methods \citep{hafner2020dreamerv2, Hafner2025, janner2019trust, clavera2020modelaugmentedactorcriticbackpropagatingpaths} offers a pathway to resolve the aforementioned issues. First, by learning dynamics from real-world robot interactions, these methods bridge the gap between simulation and reality and simplify modeling across varied environments through data acquisition rather than physics-based simulation. Furthermore, they address the sample inefficiency issue in RL via gradient signals with reduced variance \citep{clavera2020modelaugmentedactorcriticbackpropagatingpaths} relative to score-function based algorithms such as PPO \citep{schulman2017proximalpolicyoptimizationalgorithms} or SAC \citep{haarnoja2018soft, haarnoja2019softactorcriticalgorithmsapplications}. 

Instead, training agents inside learned differentiable world models using first-order gradient (FoG) methods \citep{hafner2020dreamerv2, Hafner2025, clavera2020modelaugmentedactorcriticbackpropagatingpaths} offers a promising pathway to resolve these issues. While previous FoG methods have predominantly relied on continuous online interaction within a physics simulator, applying this paradigm to learn dynamics directly from offline real-world data has the potential to effectively bridge the sim-to-real gap. Such a shift also simplifies modeling across varied environments through data acquisition rather than physics-based simulation. Furthermore, FoG methods inherently address RL sample inefficiency via gradient signals with reduced variance \citep{clavera2020modelaugmentedactorcriticbackpropagatingpaths} relative to score-function based algorithms such as PPO \citep{schulman2017proximalpolicyoptimizationalgorithms} or SAC \citep{haarnoja2018soft, haarnoja2019softactorcriticalgorithmsapplications}.

% However, the application of such methods in real-world robotics is severely constrained by the forward simulation drifts, \citep{lambert2022investigatingcompoundingpredictionerrors, xiao2019learningcombatcompoundingerrormodelbased}, and limited scalability to highly complex dynamical and visual effects common in real-world robotic settings. Recently, foundation diffusion and flow-matching world models have shown impressive capabilities for capturing such complexities in image space \cite{abc}. Nevertheless, computing gradients through such large-scale world models at the scales required for training FoG RL policies is computationally expensive and often intractable, blocking the adoption of sample-efficient first-order RL techniques.

However, the application of such methods in real-world robotics is severely constrained by forward simulation drifts \citep{lambert2022investigatingcompoundingpredictionerrors, xiao2019learningcombatcompoundingerrormodelbased}, which limit scalability to highly complex dynamical and visual effects common in real-world robotic settings. On the other hand, while recent diffusion world models have shown promising modeling capabilities \cite{dreamerv4}, computing gradients through them at the scales required for training FoG RL policies is computationally expensive and often intractable.

To address this challenge, this paper builds on Decoupled forward-backward Model-based policy Optimization (DMO) \citep{amigo2025dmo}, which separates forward simulation from backward differentiation, to make large-scale world models usable in FoG model-based RL (FoG-MBRL). Concretely, we propose a framework that maintains an offline trained large diffusion/flow-based model as a ``global'' simulator surrogate, alongside a secondary light-weight Recurrent State Space Model (RSSM)—acting as a ``local'' low-dimensional latent space world model that supplies stable, low-variance gradients without the need to backpropagate through the large global model. This decoupling resolves key obstacles to adopting FoG-MBRL alongside large-scale pixel-space world models, thereby opening the door to practical, efficient FoG-MBRL for real-world robotic problems. Through real-world experiments across a range of robotic tasks with increasing complexity, we empirically evaluate the capabilities of our approach in solving perceptual loco-manipulation tasks directly from real robot data.

\section{Related Work}
% Previous MBRL works have focused on learning from simulators, and thus aren't simulator-free like our method. It is often unclear whether these methods scale to real complex scenes. Different control strategies usually differentiate these works, with some \cite{clavera2020modelaugmentedactorcriticbackpropagatingpaths, hafner2020dreamerv2, Hafner2025, georgiev2024pwm} employing FoG-MBRL to learn a policy, others \cite{chua2018deepreinforcementlearninghandful, Hansen2022tdmpc, hansen2024tdmpc2} applying Model Predictive Control (MPC) in the world model, using zeroth-order RL algorithms \cite{janner2019trust, kurutach2018model, buckman2018sample, li2026uncertaintyawareroboticworldmodel, dreamerv4}, or using planning method \cite{khorrambakht2025worldplannermontecarlotree, schrittwieser2020mastering, niu2024lightzero, pu2024unizero}. Notable exceptions like \cite{wu2023daydreamer} or \cite{khorrambakht2025worldplannermontecarlotree} are completely simulator-free, but the first uses a variational auto-encoder, which doesn't provide the same power as the use of a diffusion model, as we do in our work, and the second one uses MPC to control, whereas we learn a policy.

\subsection{World Models and Policy Learning.}
Model-Based Reinforcement Learning (MBRL) approaches are generally distinguished by their control strategy. 
A prevalent line of work employs the learned model for online planning via Model Predictive Control (MPC) \cite{chua2018deepreinforcementlearninghandful, Hansen2022tdmpc, hansen2024tdmpc2} or Monte-Carlo Tree Search \cite{schrittwieser2020mastering, niu2024lightzero, pu2024unizero, khorrambakht2025worldplannermontecarlotree}. 
While effective, these methods incur high computational costs at inference time. 
Conversely, policy optimization methods encode control into a neural network learned inside the world model. 
However, seminal works in this category, like DreamerV3 \cite{Hafner2025}, typically rely on online interaction with a simulator. 
Simulator-free approaches like DayDreamer \cite{wu2023daydreamer} learn directly from real-world interaction, but rely on Variational Auto-Encoders (VAEs) \cite{kingma2022autoencodingvariationalbayes}, which often exhibit limited expressivity and struggle to model complex distributions with high fidelity.
In contrast, we target simulator-free learning using Diffusion Models, which offer superior modeling power but present unique policy optimization challenges.

\subsection{Gradient Estimation for Policy Learning.}
Optimization in MBRL generally falls into zeroth-order or first-order methods. 
Zeroth-order algorithms \cite{dreamerv4, janner2019trust, kurutach2018model, buckman2018sample, li2026uncertaintyawareroboticworldmodel, jordana2025introduction} treat the dynamics as a black box. 
While robust to model inaccuracies, these methods are notably sample-inefficient. 
First-order Gradient (FoG) methods \cite{clavera2020modelaugmentedactorcriticbackpropagatingpaths, georgiev2024pwm} calculate analytic gradients via backpropagation, offering high data efficiency.
However, standard FoG requires the dynamics derivatives to be computationally tractable, which excludes heavy-weight models.
A promising solution is Decoupled forward-backward Model-based policy Optimization (DMO) \cite{amigo2025dmo}, which utilizes a high-fidelity model for trajectory generation (forward) and a lightweight proxy model for gradient estimation (backward). 
While prior DMO works relied on the simulator for the forward pass, we extend this paradigm to the simulator-free setting in image space by using a ``global'' offline trained diffusion model.
This allows us to leverage high-fidelity generation without incurring the prohibitive cost of backpropagating through the heavy diffusion network.

\subsection{Diffusion for Efficient Control.}
Recent works have successfully scaled World Models using diffusion parameterizations, such as DreamerV4 \cite{dreamerv4} and DIAMOND \cite{alonso2024diffusionworldmodelingvisual}. 
However, these methods typically rely on zeroth-order optimization, which requires a large number of samples. 
We argue that the combination of heavy-weight diffusion models and sample-inefficient zeroth-order optimization is computationally expensive, as it requires generating a vast number of diffusion samples. 
Our method addresses this by employing sample-efficient FoG. 
By leveraging the decoupled architecture described above, we effectively distill the global diffusion priors into a local latent model, enabling fast analytic gradient updates and preventing the sampling bottleneck of diffusion-based zeroth-order RL methods.

\section{Method}
\begin{figure*}[t]
    \centering
    \includegraphics[width=1.0\linewidth]{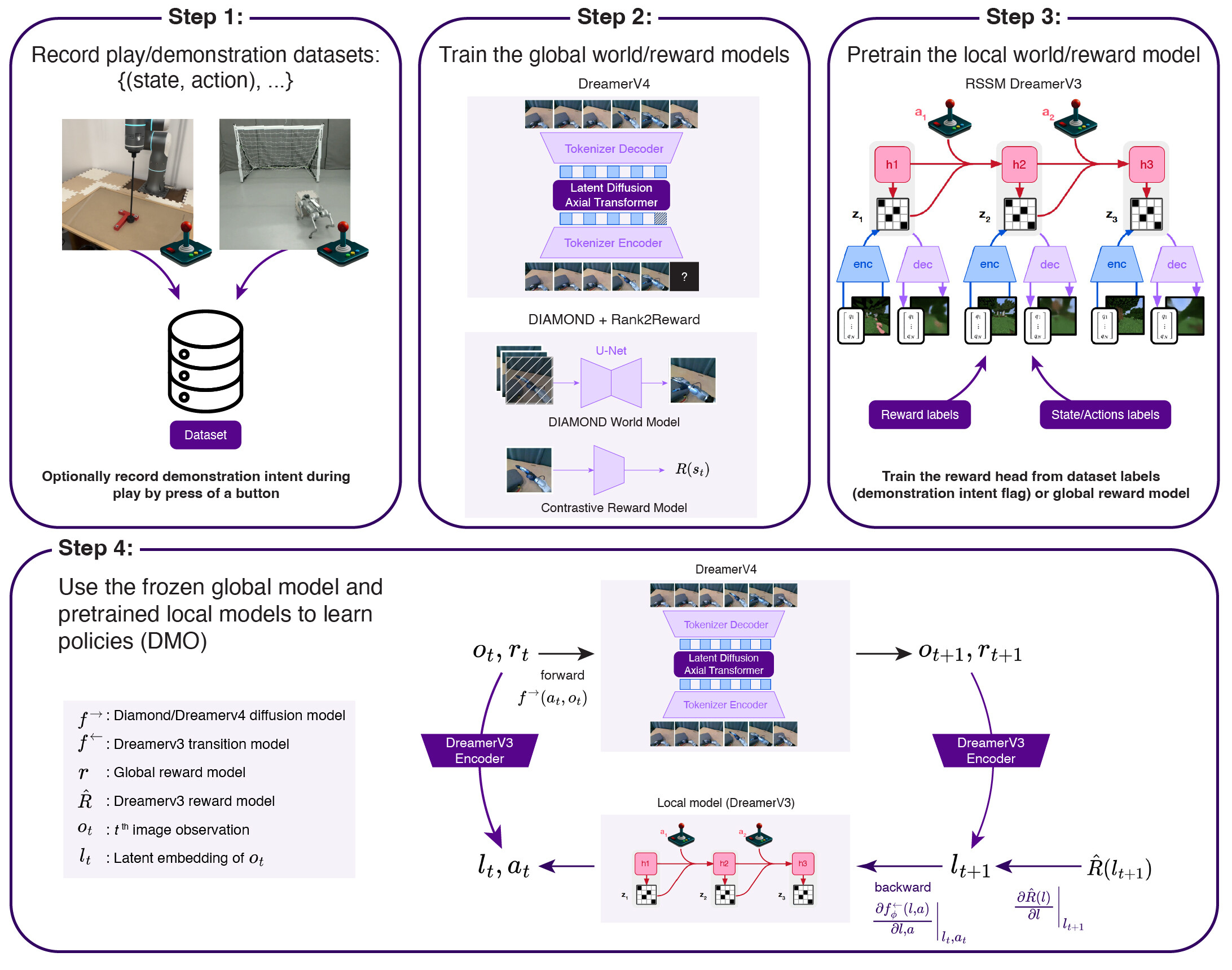}
    \caption{Overview of the proposed approach. Global world/reward models are learned from play/demonstration data (Steps 1 and 2). Then, local world/reward models are pre-trained (Step 3). Policy is optimized with the DMO algorithm (Step 4): 
    forward simulation uses the diffusion-based global world model in pixel space, trained on real robot data, to produce high-fidelity rollouts. The backward pass uses gradients computed from a local world model operating in a low-dimensional latent space. The local world/reward model is further fine-tuned during policy optimization.}
    \label{fig:dmo_diffusion_method}
    \vspace{-1.4em}
\end{figure*}
\subsection{Problem Setting and Decoupling Framework}

We consider an infinite-horizon discounted Markov Decision Process (MDP) \((\mathcal{S}, \mathcal{A}, r, \gamma, f)\), where \(\mathcal{S}\) and \(\mathcal{A}\) denote the state and action spaces, \(r: \mathcal{S} \times \mathcal{A} \to \mathbb{R}\) is the reward function, \(\gamma \in (0,1)\) is the discount factor, and \(f: \mathcal{S} \times \mathcal{A} \to \mathcal{P}(\mathcal{S})\) is the dynamics. A policy \(\pi_\theta: \mathcal{S} \to \mathcal{A}\) induces trajectories via \(a_t = \pi_\theta(s_t)\), \(s_{t+1} \sim f(s_t, a_t)\), and defines the objective
$
G(\theta) = \sum_{t=0}^{\infty} \gamma^t r(s_t, a_t).
$

First-order model-based RL (FoG-MBRL) computes policy gradients by differentiating through trajectories:
$
\nabla_\theta G(\theta)
= \sum_{t=0}^\infty \gamma^t \left(
\frac{\partial r(s_t, a_t)}{\partial s_t}\frac{d s_t}{d\theta}
+ \frac{\partial r(s_t, a_t)}{\partial a_t}\frac{d a_t}{d\theta}
\right),
$
where the state and action sensitivities evolve as
$
\frac{d s_{t+1}}{d\theta}
= \frac{\partial f(s_t,a_t)}{\partial s_t}\frac{d s_t}{d\theta}
+ \frac{\partial f(s_t,a_t)}{\partial a_t}\frac{d a_t}{d\theta}, \quad
\frac{d a_t}{d\theta}
= \frac{\partial \pi_\theta(s_t)}{\partial \theta}
+ \frac{\partial \pi_\theta(s_t)}{\partial s_t}\frac{d s_t}{d\theta}.
$
In practice, the dynamics \(f\) is often approximated by a learned model \(\hat f_\phi\), used both for rollouts \(\hat{s}_{t+1} = \hat f_\phi(\hat{s}_t, a_t)\) and for Jacobians \(\partial \hat f_\phi / \partial (s,a)\), so gradients are evaluated along model-generated trajectories \((\hat{s}_t)\) and are sensitive to compounding model error.

To address this, we adopt DMO \citep{amigo2025dmo}, which decouples forward rollouts from backward gradient computation by using distinct models: a forward model \(f^\rightarrow\) to generate high-fidelity trajectories and a backward model \(f^\leftarrow_\phi \equiv \hat{f}_\phi\) for gradient evaluation. Concretely, rather than evaluating the Jacobians of \(f^\leftarrow_\phi\) at its own next state \(s^\leftarrow_{t+1} = f^\leftarrow_\phi(s^\leftarrow_t, a_t)\), we evaluate them at the next state generated by the high-fidelity forward model \(s^\rightarrow_{t+1} = f^\rightarrow(s^\rightarrow_t, a_t)\), i.e.,
$
\left.\frac{\partial f^\leftarrow_\phi(s,a)}{\partial s}\right|_{(s^\rightarrow_{t+1}, a_{t+1})},
$
and analogously for \(\partial / \partial a\), to approximate the true dynamics derivatives in the policy gradient. This decoupling enables forward accuracy and backward tractability to be optimized independently.

% \subsection{Global Reward Models and Global Forward World Models}

% \subsubsection{Global Forward Models}
% Based on the complexity of each task and the available computational budget during each phase of our experiments, we adopt two different architectures for our global world and reward models. Specifically, we adopt a small-scale diffusion world model based on the DIAMOND \citep{alonso2024diffusionworldmodelingvisual} implementation for our preliminary experiments on the canonical Push-T setup. Thanks to the optimized noise scheduling and pre-/post-conditioning schemes used in this architecture \citep{edm} DIAMOND achieves efficient interactive inference for long rollout durations without considerable degradation. However, for our ego-centric tasks where a long temporal context length is essential for maintaining observability (e.g. object occlusions during manipulation), we implement a transformer-based latent diffusion world model, based on the recent DreamerV4 \citep{dreamerv4} architecture. This architecture is suitable for our task because it enables fast inference through shortcut learning \citep{shortcut-learning} (few diffusion steps) and causal diffusion forcing \citep{diff_forcing} (KV-cache-friendly real-time recurrent generation). 

\subsection{Global World Model}
\label{sec:method-global-world}
Based on the complexity of each task and the available computational budget during each phase of our experiments, we adopt two different architectures for our global world model. For preliminary experiments on a canonical Push-T setup, we use a small-scale (4M parameters) diffusion-based world model based on the DIAMOND \citep{alonso2024diffusionworldmodelingvisual} architecture due to long horizon prediction accuracy and efficient inference speeds. 
% Thanks to the optimized noise scheduling and pre-/post-conditioning schemes used in this architecture \citep{edm}, DIAMOND achieves efficient interactive inference for long rollout durations without considerable degradation.
For all other tasks, we implemented a transformer-based latent diffusion world model, based on the recent DreamerV4 \citep{dreamerv4} architecture. We ensure efficient inference by adopting shortcut learning \citep{shortcut-learning} (few diffusion steps) and causal diffusion forcing \citep{diff_forcing} (KV-cache-friendly real-time recurrent generation). We also integrate proprioceptive data as an additional modality to the tokenizer for our Humanoid G1 task. 

\subsection{Global Reward Model}
\label{sec:method-global-reward}
Learning in pixel space requires image-based reward functions. We adopt two reward modeling strategies depending on the underlying global world model. For the Push-T experiments with the DIAMOND world model, we train a contrastive reward model on random play and, optionally, demonstration videos, following \citep{rank2reward}. 
% Using passive video data (play and optional demonstrations without actions),
Specifically, we learn a goal-conditioned energy function \(f_\psi(s_t \mid s_{\text{start}}, s_{\text{goal}})\) that assigns higher scores to states temporally closer to the goal. 
%Training uses a Bradley--Terry objective over temporally ordered frame pairs:
% \begin{equation}
% \label{eq:rank2reward}
% \mathcal{L}(\psi)
% = \frac{e^{f_\psi(s_i)}}{e^{f_\psi(s_i)} + e^{f_\psi(s_j)}} \mathbf{1}\{ i > j \}
% + \frac{e^{f_\psi(s_j)}}{e^{f_\psi(s_i)} + e^{f_\psi(s_j)}} \mathbf{1}\{ i \le j \},
% \end{equation}
% where \(s_i, s_j\) are sampled between \(s_{\text{start}}\) and \(s_{\text{goal}}\).

For experiments using our transformer-based world model, we instead train a lightweight reward head that classifies whether a frame belongs to the task demonstration distribution; on Push Cube, frames where the cube reaches the goal carry a higher label (10). Labels are provided by a human to indicate task-relevant behavior, either during data collection (e.g., button presses during play) or in a post-processing pass. 
% Throughout the paper, we refer to these as \emph{global reward models}, in contrast to the \emph{local reward models} used for the backward pass. 
% Task-specific instantiations are detailed in Sec.~\ref{sec:exp-world-reward}.

% \subsection{Local Reward Model and Local Backward World Model}
% For the backward model \(f^\leftarrow_\phi\), we adopt the recurrent state-space model (RSSM) architecture adopted in DreamerV3 \citep{Hafner2025}, which acts as a joint local latent dynamics and reward model. The model is first pretrained offline on the play dataset, with the global reward model/head providing the reward ground truth. During online RL training, we continue fine-tuning \(f^\leftarrow_\phi\) and its reward head using trajectories generated by the forward global model \(f^\rightarrow\), ensuring that the RSSM maintains high local accuracy with respect to the current policy's conditional action distribution. To avoid differentiating through pixels (for memory efficiency), an image encoder projects observations into a compact latent space, where \(f^\leftarrow_\phi\) operates exclusively; notably, the policy gradients do not flow through the encoder. In the DMO framework, unlike DreamerV3 or other MBRL frameworks, the transition model in \(f^\leftarrow_\phi\) requires only single-step accuracy, as Jacobians are evaluated at the encodings of precise forward-simulated images from the global model \(f^\rightarrow\).

\subsection{Local Reward and World Models}
\label{sec:method-local}
For the backward model \(f^\leftarrow_\phi\), we adopt the recurrent state-space model (RSSM) architecture used in DreamerV3 \citep{Hafner2025}, which acts as a joint local latent dynamics and reward model. The model is first pretrained offline on the play dataset. Its reward targets are the predictions of the global reward model on Push-T and, on Push Cube and G1, the dataset's per-frame reward labels, which also supervise the global reward head. During online RL training, we continue fine-tuning \(f^\leftarrow_\phi\) and its reward head using trajectories generated by the forward global model \(f^\rightarrow\), with the global reward model's predictions as reward targets, ensuring that the RSSM maintains high local accuracy with respect to the current policy's conditional action distribution. To avoid differentiating through pixels (for memory efficiency), an image encoder projects observations into a compact latent space, where \(f^\leftarrow_\phi\) operates exclusively (the policy gradients do not flow through the encoder). In the DMO framework, unlike DreamerV3 or other MBRL frameworks, the transition model \(f^\leftarrow_\phi\) needs to be accurate only over single forward steps, as Jacobians are evaluated at the encodings of precise forward-simulated images from the global model \(f^\rightarrow\).

\paragraph{Relation to DMO.} DMO cannot be applied as is in our setting. It evaluates the backward model at the states produced by the forward model, so both must share one state space. In image-space problems, a high-fidelity forward model has to operate on high-dimensional states: on Push Cube, our global model conditions on up to 96 frames of 256 latent tokens of width 16, against the 68-dimensional state of DMO's quadruped task, an increase of more than $5{,}700\times$. Giving the local model this dimensionality would make the backward pass prohibitively expensive. We therefore couple the two models through image space: each frame generated by \(f^\rightarrow\) (decoded from its latent by the tokenizer decoder for the transformer-based model) is re-encoded into the local latent by the RSSM's image encoder, both without gradient. The Jacobians of the global model, taken in its own latent space, are then approximated by those of the local model, taken in its own, so the two latent spaces need not have the same dimension. Two further components that DMO obtained from its simulator must be learned here. First, DMO's backward model operated on the simulator's ego-body state, for which rewards can be written by hand, whereas we must learn the reward from images (Sec.~\ref{sec:method-global-reward}). Second, a simulator is exact by construction, whereas our world model, learned from offline data, is never re-grounded on real observations during episode-long rollouts and may hallucinate; we address this with the patching procedure of Sec.~\ref{sec:data-protocol}.

\subsection{Data Collection and Training Protocol}
\label{sec:data-protocol}
% We collect an unstructured high-entropy action-image pair \emph{play dataset} on the real robot to train our global world model from scratch (reducing data need through foundation pretrained world models is the topic of future research). Although high-entropy interactions can degrade performance in Behavioral Cloning (BC) methods \cite{zhu2025should}, they contain valuable interaction and dynamics primitives suitable for training world models that are suitable for RL training. In fact, these broader interactions form a superset that encompasses more targeted, narrow task-demonstration skills \cite{khorrambakht2025worldplannermontecarlotree}.. Furthermore, such data does not require clean temporal segmentation of the sequence into clearly defined tasks and does not mandate frequent environment resets during data collection, making it less expensive and easier to collect and reuse from other tasks.

We collect an unstructured high-entropy action-image pair \emph{play dataset} on the real robot to train our global world model from scratch, except for the Push Cube and G1 tokenizers, which are initialized from tokenizers pretrained on the SOAR robot dataset~\cite{zhou2024soar} (reducing data needs through foundation-pretrained world models is the topic of future research). Such data does not require clean temporal segmentation of the sequence into clearly defined tasks and does not mandate frequent environment resets during data collection, making it less expensive and easier to collect and reuse from other tasks.

To ensure that the global world model accurately captures both the large workspace of the Go2 Push Cube task and the fine-grained contact dynamics between the robot head and the cube, as well as the complex multi-contact interactions involved in the high-DoF Humanoid G1 Grab Box task, we use two complementary refinement procedures. First, we qualitatively probe the world model through immersive exploration, using a Meta Quest 3 headset for Go2 Push Cube and a Pico headset with IMUs for Humanoid G1, to identify poorly modeled regions of the state space; we then collect additional data in those regions and fine-tune the model accordingly. This process is feasible because the world-model action space is expressed in terms of high-level actions that a human operator can reproduce. Second, we use RL to discover policies that exploit inaccuracies in either the world model or the reward model to obtain artificially high returns. We then deploy these exploiting policies on the real robot to gather targeted data and fine-tune the model, thereby patching the weak regions that enabled the exploit. On G1, we ran seven rounds of such targeted data collection and fine-tuning.

\section{Experiments}
\begin{figure}[t]
    \centering
    \includegraphics[width=\linewidth]{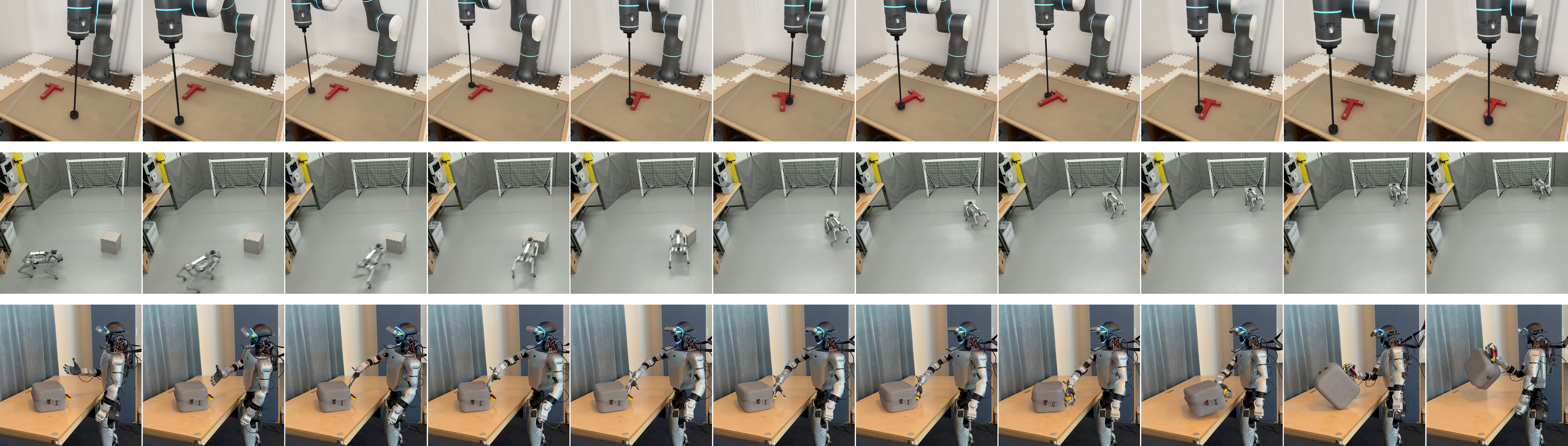}
    \caption{Real-world zero-shot deployment of our DMO policies trained in the world model.}
    \label{fig:real-world-deployment}
    \vspace{-1em}
\end{figure}
\subsection{Tasks and Setups}
\label{sec:tasks-and-setups}
% In this section, we describe the tasks and robots considered in this study. Importantly, we favor fewer but more diverse real-world deployments over a large number of simulated experiments to maintain focus on real-world modeling and transferability challenges that world models aim to solve.
To truly evaluate the potential of our approach to address the unmodeled complexities of real-world settings and its zero-shot sim2real transferability, we favor fewer, more representative real-world deployments over a large number of simulated tests that fall short of capturing the modeling complexities of real robotics tasks. Specifically, we consider three carefully chosen tasks and robot embodiments as introduced in the next subsections. It is important to note that, unlike other RL finetuning approaches that use world models \cite{zhang2026learningvisualfeaturebasedworld, sharma2026worldgymnasttrainingrobotsreinforcement}, we train our policies entirely within the world model from scratch, without any behavior-cloning guidance or regularization. 

\subsubsection{Push-T}
We adopt the Push-T canonical task as an easier-to-model but non-trivial motion planning problem. Specifically, this task requires pushing a T-shaped object from a random pose on a flat 2D surface to the center of the board in a predefined orientation. The action is the tool's vertical and horizontal velocity, and stacks of 4 consecutive frames are used as states, both provided at 5 Hz. We use a 7-DoF Flexiv Rizon-10S robot with an onboard task-space velocity controller to build our environment. We adopt a small 4M-parameter DIAMOND world model as the global forward model \(f^\rightarrow\) and train it on 4 hours of play data.

\subsubsection{Ego-Centric Push Cube}
In this task, we increase the difficulty by introducing partial observability, walking dynamics, and higher visual complexity. Specifically, a Unitree Go2 robot with ego-centric vision is tasked with pushing a cubical object into a soccer goal using its body. Body velocity commands to a low-level RL locomotion policy are taken as action and provided at 5 Hz. This task requires the policy and the world model to be robust to partial occlusions of the view by the object, maintain a short-term memory of where it was when it falls out of view, and handle the geometric complexity of a large room in image space. Thus, we train a 1.0B-parameter implementation of our transformer-based latent diffusion world model on 12 hours of data to serve as the forward model \(f^\rightarrow\) with a reward head. 

\begin{figure}[t]
    \centering
    \includegraphics[width=\linewidth]{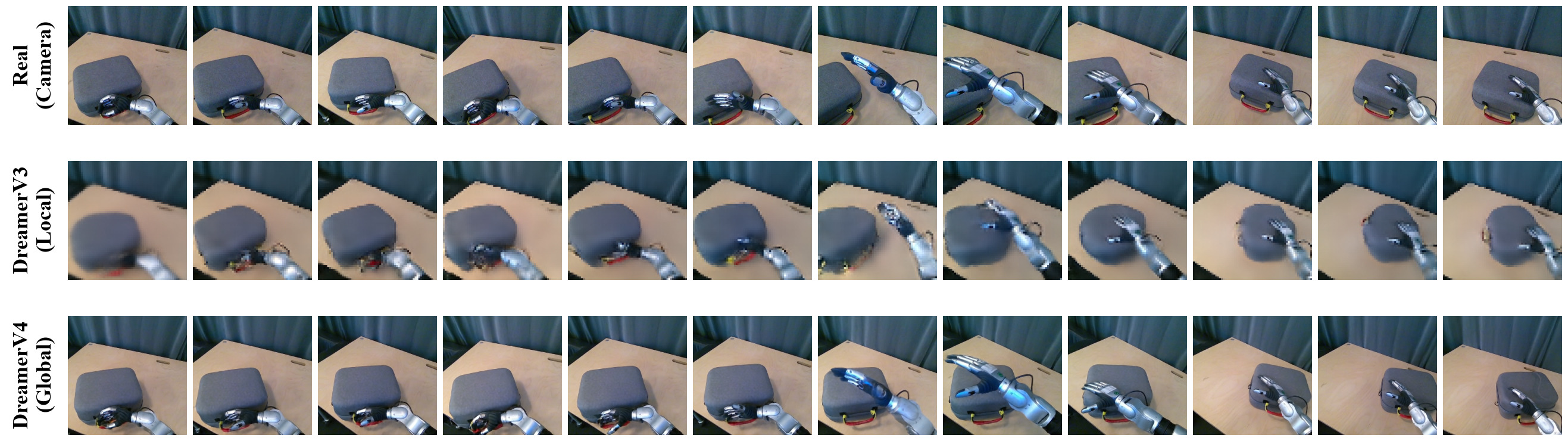}
    \caption{Unrolling of real and model-predicted trajectories on the G1 humanoid manipulation task. We compare the real camera observation (\textbf{top}), the local DreamerV3 RSSM trajectory (\textbf{middle}), and the global DreamerV4 diffusion trajectory (\textbf{bottom}) at time steps \(0, 5, 10, \dots, 55\) (\(5.5\) s at \(10\) Hz), with both models initialized from the first real frame. The global diffusion model maintains high visual fidelity and accurate contact dynamics over the long horizon.}
    \label{fig:g1_compare_dreamer_diffusion_main}
    \vspace{-1em}
\end{figure}

\subsubsection{Humanoid Grasp and Lift}
Finally, we extend the previous experiments by considering grasping skill and a high-DoF Unitree G1 humanoid robot equipped with a BrainCo dexterous hand. The task is to grasp a box through its flexible handle and lift it above the table. %In this task, the robot needs to reason about fine hand-pose adjustments to approach the handle carefully, close the hand at the right time, and lift the box in a stable manner.
The world model is conditioned on 22-dimensional commands: 17 upper-body joint-position targets (3 waist and 14 arm joints), 2 hand-grip commands, and 3 lower-body locomotion commands (base velocities and yaw rate). The policy outputs the 19 upper-body commands through bounded per-step increments (at most $0.02$~rad per step for the joints); on the real robot, the left arm is kept fixed and its commands are ignored, so the deployed policy controls 11 DoF: the waist, the right arm and the right hand. The humanoid is controlled by the NVIDIA SONIC \cite{luo2025sonic} full-body tracking RL policy at 50 Hz, and receives body-tracking commands from our policy at 10 Hz. We use the same type of global forward and reward model as for Push Cube, at a larger scale (1.4B parameters, $256\times256$ images; Table~\ref{tab:hyperparameters}), trained on 20.9 hours of data: 16.5 hours of play data and 4.4 hours collected during the seven rounds of targeted data collection (Sec.~\ref{sec:data-protocol}). The dynamics model is trained from scratch, and the tokenizer is initialized from a tokenizer pretrained on the SOAR robot dataset~\cite{zhou2024soar}.

\subsection{Baselines and Ablations}
We compare our method against the following baselines. (1) Model-free RL trained inside the same global world model, with the same reward model and the same initial-state distribution as our method: PPO \cite{schulman2017proximalpolicyoptimizationalgorithms} and SAC \cite{haarnoja2018soft}, both trained directly on image observations. (2) Offline RL: IQL \cite{kostrikov2022iql}, trained on the recorded data of each task without any interaction, and shown at its best evaluation in the global world model. (3) Behavior Cloning (BC): the Action Chunking Transformer (ACT) \cite{zhao2023learningfinegrainedbimanualmanipulation}, a well-known representative of BC policies, on Push Cube and G1 Grab Box.\footnote{We decided to include a BC baseline after the Push-T setup had been dismantled, so Push-T has no BC result.} SAC and IQL are evaluated in the global world model only (Fig.~\ref{fig:aggregated_efficiency}), while PPO, ACT and our method are deployed on the real robots (Tables~\ref{tab:traj_metrics}--\ref{tab:success-rates-g1}).
%It is important to note that, unlike common RL finetuning approaches that use world models \cite{zhang2026learningvisualfeaturebasedworld, sharma2026worldgymnasttrainingrobotsreinforcement}, we train our policies entirely within the world model from scratch, without any behavior-cloning guidance or regularization. 

%Furthermore, to our knowledge, existing FoG-MBRL methods, such as the original DreamerV3 algorithm, are not suitable for working with the diffusion-based large world models adopted in this paper, for the same gradient computation challenges for which we proposed our method. As a result, we do not include this family of works among our baselines. Nevertheless, we do test whether a DreamerV3-only world model would suffice to learn a good policy in a baseline named ``No Diffusion''.

We also include two ablations: the first, ``No Diffusion'', replaces the global diffusion world model (DIAMOND or DreamerV4) for the forward pass with the DreamerV3 RSSM model and thus uses the same RSSM model for the forward and backward passes. The aim is to demonstrate that powerful models are indeed needed for the forward pass to accurately capture the complexity of real-world robotics scenarios. The second ablation, ``No RSSM Finetuning'', evaluates the need to finetune the RSSM during RL training, providing the practitioner with additional insights into our design decisions.

%, existing MBRL methods do not support image-space policy learning without access to a simulator. Standard Model-Based RL baselines, such as the original DreamerV3, typically rely on online interaction with a simulator.Nevertheless, we demonstrate the infeasibility of removing the simulator (relying on DreamerV3 as a differentiable simulator) under a baseline called ``No Diffusion''. As expected and shown in Fig.~\ref{fig:combined_compare_dreamer_diffusion}, this leads to poor forward simulation quality and thereby converging to unsuccessful policies (Tab.~\ref{tab:success-rates}). Here, it is important to note that removing the DreamerV3 and relying solely on direct backpropagation through the global model is also infeasible due to the model's much larger scale and the deep computation graph imposed by the multi-step diffusion process. 

%Figure \ref{fig:compare_dreamer_diffusion} and Figure \ref{fig:go2_compare_dreamer_diffusion} show that DreamerV3 generalizes less effectively than the diffusion world model. %However, its short-horizon accuracy around the current policy remains sufficient to provide meaningful Jacobians. As a result, our proposed decoupled design, using the local model for tractable gradient computation and, by decoupling, exploiting the diffusion model’s broad generalization, delivers significant improvements in both sample and time efficiency.

\subsection{Results}\label{sec:results}
\begin{figure}[t]
    \centering
    \includegraphics[width=0.9\linewidth]{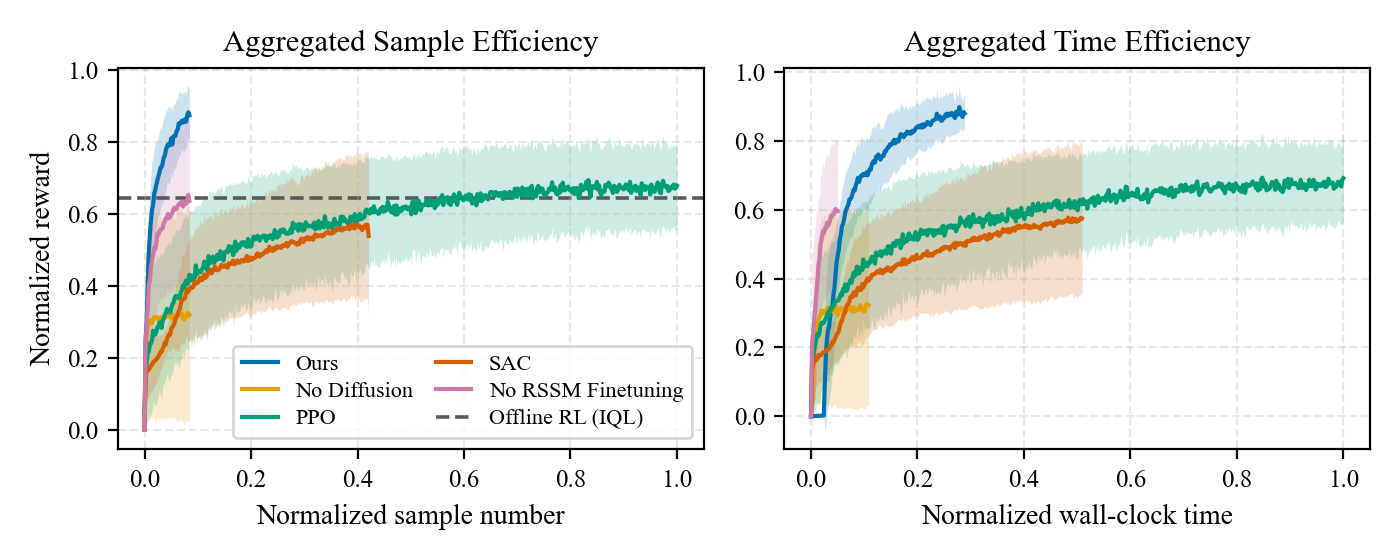}
    \caption{Aggregated performance metrics across all evaluation tasks. \textbf{Left:} Normalized reward relative to normalized sample number; offline RL (IQL) uses no environment samples and is shown as a horizontal line at its best evaluation. \textbf{Right:} Normalized reward relative to normalized wall-clock time. Every online experiment was run with 4 seeds.}
    \label{fig:aggregated_efficiency}
\end{figure}
% \begin{figure}[h]
%     \centering
%     \centering
%     \includegraphics[width=1.0\linewidth]{paper-template-latex/Assets/sample_efficiency_pusht_new.png}
%     \caption{Efficiency comparison. Left: Sample efficiency on the PushT task for DMO (8M samples) versus PPO (40M samples) and the No Diffusion ablation. Right: Time efficiency comparison between DMO and PPO and the No Diffusion ablation.}
%     \label{fig:efficiency_pusht}
% \end{figure}
% \begin{figure}[h]
%     \centering
%     \includegraphics[width=1.0\linewidth]{paper-template-latex/Assets/sample_time_efficiency_push_cube.png}
%     \caption{Efficiency comparison. Left: Sample efficiency on the Push Cube task for DMO (4M samples) versus PPO (25M samples) and the No Diffusion ablation. Right: Time efficiency comparison between DMO and PPO.}
%     \label{fig:sample_time_efficiency_push_cube}
% \end{figure}

% In the preamble

The results in Fig.~\ref{fig:aggregated_efficiency} demonstrate that DMO achieves superior sample and time efficiency compared to PPO and SAC across all tasks, and reaches a higher reward than the best evaluation of offline RL (IQL). Notably, the time efficiency gains shown in Fig.~\ref{fig:aggregated_efficiency} validate our hypothesis that first-order policy gradients are more suitable than zeroth-order variants for computationally heavy world models (DMO vs PPO and SAC). Refer to Appendix~\ref{apx:per_task_efficiency} for per-task efficiency results.

Furthermore, the poor performance of the ``No Diffusion'' ablation in Fig.~\ref{fig:aggregated_efficiency} further underscores the necessity of a high-fidelity global world model for forward simulation of more complex tasks. As shown in Fig.~\ref{fig:g1_compare_dreamer_diffusion_main}, and in Appendix~\ref{apx:world_model_rollouts}, DreamerV3 generalizes less effectively than the diffusion world model. %However, its short-horizon accuracy around the current policy remains sufficient to provide meaningful Jacobians for policy optimization as the DMO policy optimization result show. %As a result, our proposed decoupled design, using the local model for tractable gradient computation and, by decoupling, exploiting the diffusion model’s broad generalization, delivers significant improvements in both sample and time efficiency.
Nevertheless, using the local model for tractable gradient computation and, by decoupling, leveraging the diffusion model’s broad generalization, delivers significant improvements in both sample and time efficiency.
%
% \begin{table}[b]
% \centering
% \caption{Success Rates Across Algorithms for Push-T (out of 10 Tries)}
% \label{tab:success-rates}
% \begin{tabular}{lc}
% \toprule
% Algorithm & Successes / 10 \\
% \midrule
% DMO & 9/10 \\
% PPO & 1/10\footnotemark  \\
% No Diffusion & 0/10 \\
% \bottomrule
% \end{tabular}
% \end{table}
%

\begin{table}[h]
    \centering
    \caption{Trajectory Quality Analysis. We report the mean$\pm$std over $N=10$ randomly sampled evaluation episodes on the real robot for the Push Cube task. \emph{Success Rate}: fraction of episodes reaching the goal; \emph{Steps}: lower is better; \emph{Straightness}: displacement/path length, $\uparrow$ is straighter; \emph{Curvature}: yaw change per meter, $\downarrow$ is smoother. Metrics computed over successful episodes only. $^{\dagger}$For the BC baseline, success is defined as guiding the cube to the immediate pre-goal vicinity, as the cube never fully enters the goal.}
    \label{tab:traj_metrics}
    \vspace{0.5em}
    \resizebox{\linewidth}{!}{%
    \begin{tabular}{lcccc}
    \toprule
    Method & Success Rate ($\uparrow$) & Steps to Success ($\downarrow$) & Straightness Index ($\uparrow$) & Curvature (rad/m) ($\downarrow$) \\
    \midrule
    DMO (Ours)   & \textbf{90\%} & 177.2 $\pm$ 52.2 & \textbf{0.520 $\pm$ 0.266} & \textbf{0.692 $\pm$ 0.206} \\
    BC (ACT)$^{\dagger}$ & 60\% & 371.0 $\pm$ 94.0 & 0.330 $\pm$ 0.147 & 0.761 $\pm$ 0.151 \\
    PPO          & 40\%          & 173.5 $\pm$ 20.3 & 0.233 $\pm$ 0.182 & 1.222 $\pm$ 0.306 \\
    No Diffusion & 0\%           & --               & --                & --                \\
    \bottomrule
    \end{tabular}%
    }
\end{table}

\begin{table}[h]
    \centering
    % Left side: Push-T Success Rates Table
    \begin{minipage}[t]{0.48\linewidth}
        \centering
        \caption{Success Rates Across Algorithms for Push-T (on the real robot, out of 10 Tries). $^{\dagger}$Under a relaxed success criterion (counting runs where the T briefly passes through the correct pose and position without stopping), PPO achieves 4/10 successes.}
        \label{tab:success-rates-pusht}
        \vspace{0.5em}
        \begin{tabular}{lc}
        \toprule
        Algorithm & Successes / 10 \\
        \midrule
        DMO (Ours)   & 9/10 \\
        PPO          & 1/10$^{\dagger}$ \\
        No Diffusion & 0/10 \\
        \bottomrule
        \end{tabular}
    \end{minipage}%
    \hfill%
    % Right side: G1 Success Rates Table
    \begin{minipage}[t]{0.48\linewidth}
        \centering
        \caption{Success Rates Across Algorithms for G1 Grab Box (on the real robot, out of 10 Tries).}
        \label{tab:success-rates-g1}
        \vspace{0.5em}
        \begin{tabular}{lc}
        \toprule
        Algorithm & Successes / 10 \\
        \midrule
        DMO (Ours)   & 8/10 \\
        BC (ACT)     & 7/10 \\
        PPO          & 1/10 \\
        No Diffusion & 0/10 \\
        \bottomrule
        \end{tabular}
    \end{minipage}
\end{table}

Table~\ref{tab:success-rates-pusht} reports the success rates for the Push-T real-robot experiment (Fig.~\ref{fig:pusht-deployment}) and confirms the advantage of our approach compared to the baselines. The ``No Diffusion'' ablation did not lead to the successful achievement of the task, and PPO only managed to solve the task once. We observe similar patterns in all other tasks, as shown in Tables~\ref{tab:traj_metrics} and~\ref{tab:success-rates-g1}.

% To gain deeper insight beyond success-rate statistics, we conduct a quantitative study that investigates various performance axes of each baseline for the ego-centric push-cube task. 

It is worth noting that for the Push Cube task, the episode length had to be increased to 128 steps to train the PPO policy deployed on the robot (as opposed to 64 for DMO and for the learning curves of Fig.~\ref{fig:aggregated_efficiency}); with 64-step episodes, PPO fell into a suboptimal local solution where it repeatedly approaches and avoids the cube to collect rewards of 1. In the meantime, DMO succeeded in both settings and demonstrates robustness to hyperparameter choices.

Regarding the ACT baselines, we observed common BC problems such as not seeking long-term rewards (e.g., stopping in front of the goal in the Push Cube task), not discovering solutions or failing at poorly covered regions, and being sensitive to the consistency of the demonstrations (demonstrator's task-solving strategy).

Furthermore, the Push Cube experiment revealed distinct behavioral differences across methods, particularly in object localization, task completion, and control smoothness. In Table~\ref{tab:traj_metrics} we report \emph{Straightness Index} (the ratio of Euclidean displacement to total path length, where $1.0$ is optimal) and \emph{Curvature} (average yaw change per meter) that verify this observation. 

Finally, Fig.~\ref{fig:real-world-deployment} shows snapshots of the deployed DMO policy across tasks. For more qualitative demonstrations, refer to the appendix.
%\footnotetext[2]{ACT never reaches the terminal state and gets stuck at the goal entrance, and thus we cannot report on steps to success for it.}

\section{Limitations}
In this paper, we train our global world models from scratch (except for the Push Cube and G1 tokenizers, initialized from tokenizers pretrained on SOAR robot data) and assume hand-labeled rewards for the reward head (Go2 Push Cube and G1 Grab Box tasks). Consequently, our play data must cover a wide range of physical interactions in image space, which challenges its adoption for more complex and long-horizon tasks. Our future work will focus on leveraging prior knowledge from internet-scale pre-trained video models to alleviate this limitation. Furthermore, the automated exploration and data collection scheme briefly explored under this project introduces a promising orthogonal solution worth further future investigation. This argument also applies to reward modeling, where leveraging VLM-based prior knowledge for reward labeling offers a promising solution for automating additional components within our system.

\section{Conclusion}
In summary, our framework bridges the limitations of simulator-centric RL by learning diffusion world models from real data, yielding visually realistic simulations that substantially reduce sim-to-real discrepancies in manipulation tasks. The decoupling mechanism leverages the comparative strengths of models—global diffusion for accurate, generalizable forward predictions and local latent models for tractable backward gradients—maintaining efficiency despite the high dimensionality of pixel inputs. Notably, our method enables learning policies directly from real robot data, using images as input, and transfers zero-shot with a high success rate on the real robot. Importantly, our experiments demonstrated practical improvements in task performance over pure behavior cloning. Our method provides a safe proxy for real-robot training, allowing scalable policy optimization without risking hardware damage or unsafe behaviors, paving the way for robust vision-based robotics in unstructured environments.

\clearpage
% The acknowledgments are automatically included only in the final and preprint versions of the paper.
\acknowledgments{This work was granted access to the HPC resources of IDRIS under the allocations AD011015316R1, A0201017555 and A0191016928 made by GENCI. This work was in part supported by the US National Science Foundation grants 2026479, 2222815, and 2315396, and the French AI Interdisciplinary Institute ANITI. ANITI is funded by the France 2030 program (grant agreement ANR-23-IACL-0002). It was also supported in part by the AGIMUS project, funded by the European Union under GA no.~101070165, and the ANR NERL project (grant ANR-23-CE94-0004).}

% no \bibliographystyle is required, since the corl style is automatically used.
\bibliography{example}
\clearpage

\appendix
\section{Appendix}
\subsection{Qualitative Comparison of World Model Rollouts}
\label{apx:world_model_rollouts}

\begin{figure}[H]
    \centering
    \begin{subfigure}{\linewidth}
        \centering
        \includegraphics[width=\linewidth]{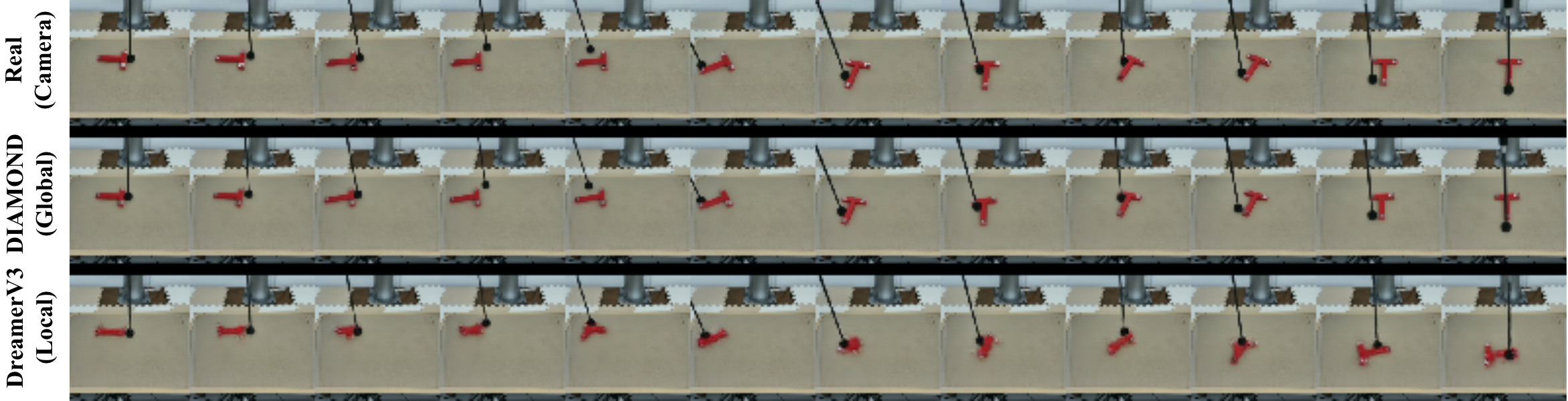}
        \caption{}
        \label{fig:compare_dreamer_diffusion}
    \end{subfigure}

    %\vspace{0.5em}

    \begin{subfigure}{\linewidth}
        \centering
        \includegraphics[width=\linewidth]{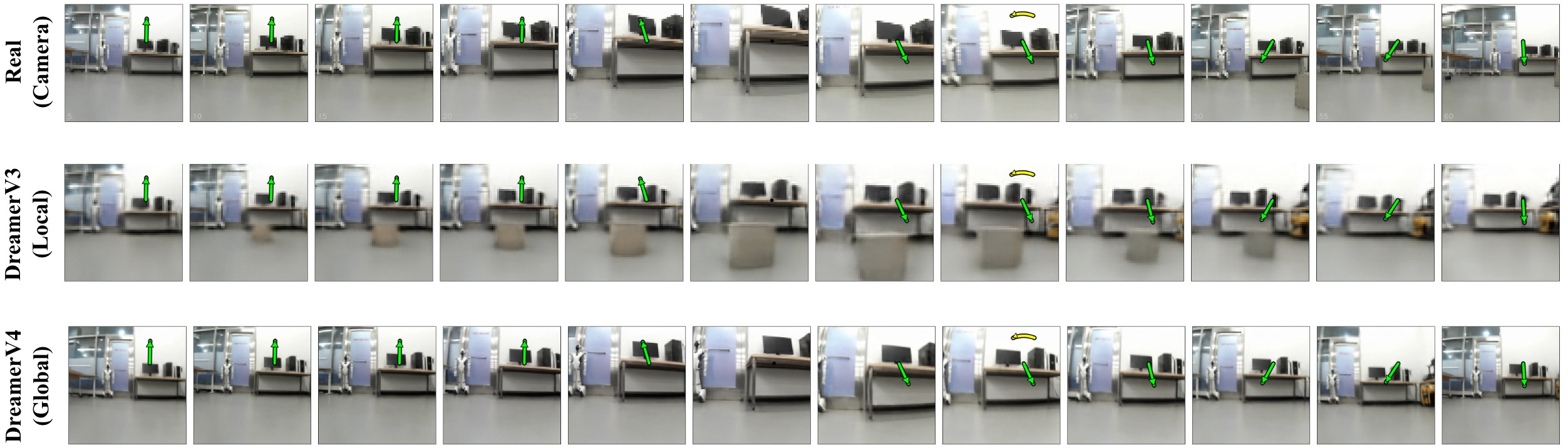}
        \caption{}
        \label{fig:go2_compare_dreamer_diffusion}
    \end{subfigure}

    \caption{(a) Unrolling of real and model-predicted trajectories comparing the real trajectory (top), the DIAMOND diffusion trajectory (middle), and the DreamerV3 trajectory (bottom), every 5 steps (\(12\) s). (b) Unrolling of real and model-predicted trajectories comparing the real trajectory (top), the DreamerV3 trajectory (middle), and the DreamerV4 diffusion trajectory (bottom), every 5 steps (\(12\) s), with both models initialized with zero context, and the cube initially occluded. Note that the local model violates object permanence by spawning the cube throughout the rollout.}
    \label{fig:combined_compare_dreamer_diffusion}
\end{figure}

\subsection{Humanoid G1 Experiments: World Model Exploit Patching}
\label{sec:appendix_g1_experiments}

In this section, we provide additional visual details regarding the world model patching process evaluated on the Unitree G1 humanoid robot. The task involves the robot using a dexterous hand to grasp and lift a box equipped with a semi-soft handle. 

As discussed in the main text, the reinforcement learning (RL) policy initially discovered exploits within the dynamics of the original world model. To address this, we deployed the exploiting policy directly on the real robot to collect targeted patching data during these exploitation attempts. This real-world data was then utilized to patch and refine the world model. We ran seven such rounds, which added 0.19, 0.55, 0.74, 0.80, 1.00, 0.72 and 0.43 hours of data (4.4 hours in total). After each round, the dynamics model pretrained on the 16.5 hours of play data was fine-tuned for 3k updates, with a newly initialized reward head, on the play data and all the data collected so far.

\begin{figure}[H]
    \centering
    % Ensure the image file is in the same directory or provide the correct path (e.g., figures/exploit_comparisons_grid.jpg)
    \includegraphics[width=\textwidth]{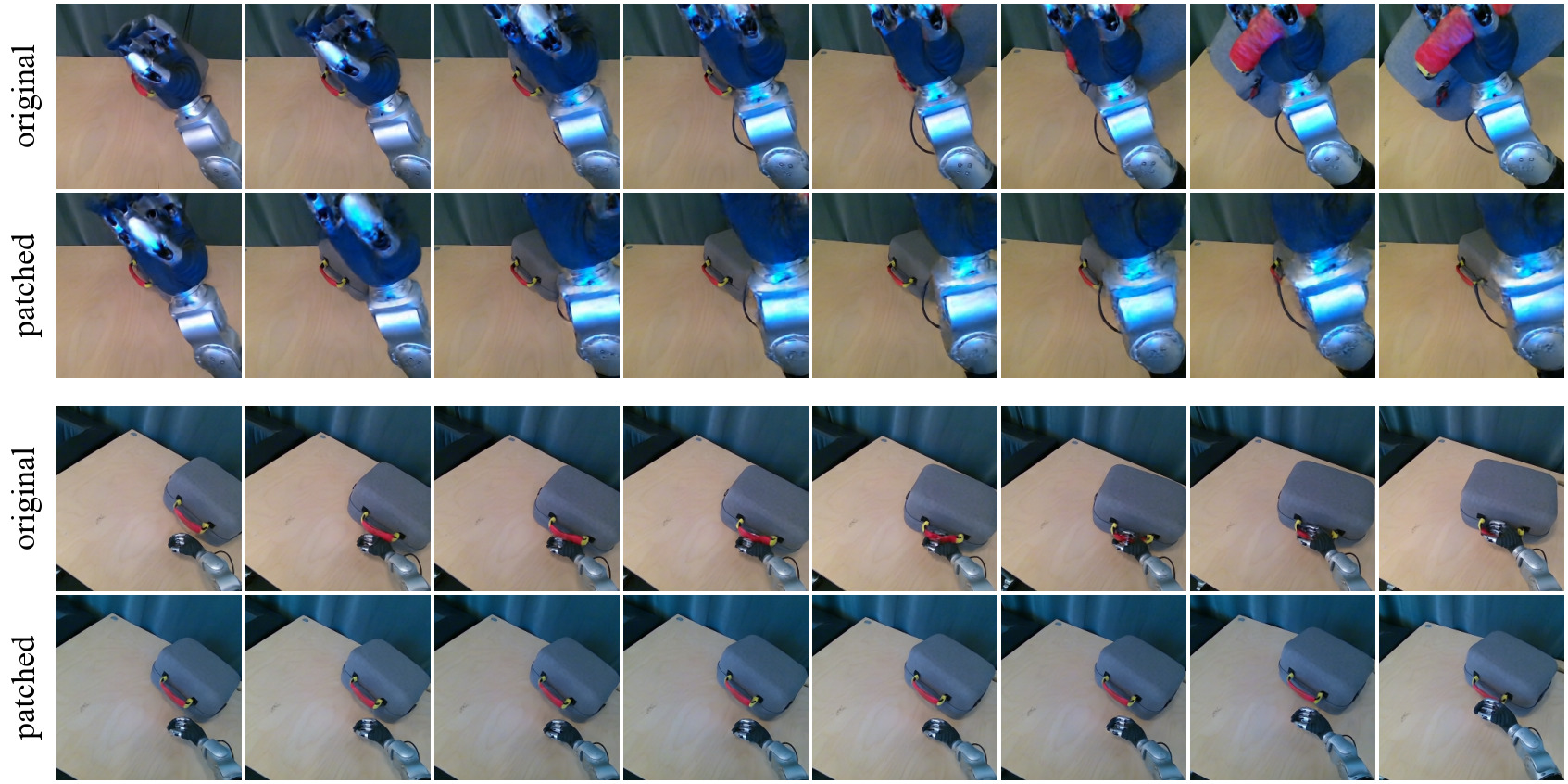}
    \caption{Visual comparison of the world model before and after patching on the Unitree G1 Grab Box task. Rows labeled \textit{original} display the unpatched world model where the RL policy discovered physical exploits to artificially maximize reward. In the first rollout (\textbf{top two rows}), the policy exploits the model by moving the hand into a lifting pose, causing the box to spontaneously teleport into its grasp. In the second rollout (\textbf{bottom two rows}), the hand remains distant but pulls the box toward itself using an ``invisible force.'' Rows labeled \textit{patched} show the corrected world model rollouts after being fine-tuned with real-world data collected by deploying these exploiting policies, effectively eliminating the physical hallucinations.}
    \label{fig:exploit_comparisons}
\end{figure}

Figure \ref{fig:exploit_comparisons} illustrates the rollouts of the world model before and after this patching process, demonstrating how the updated model successfully corrects the physical inconsistencies that were previously exploited by the RL policy.

\subsection{Humanoid G1: World-Model Accuracy, Gradient Alignment and Compute Cost}
\label{apx:g1_analysis}

\paragraph{World-model accuracy and uncertainty.} On G1, we unroll the global world model and the local RSSM open-loop over the full 64-step episode on 64 held-out trajectories, with the context prefilled with 32 real frames and $K=8$ samples per trajectory. The global model reaches an LPIPS~\cite{zhang2018unreasonable} of 0.295 against 0.425 for the RSSM (lower is better), and a spread-to-error ratio of 0.665 against 0.275 (higher is better). The RSSM is confidently wrong, consistent with the failure of the ``No Diffusion'' ablation on all three real-robot tasks (Tables~\ref{tab:traj_metrics}--\ref{tab:success-rates-g1}).

\paragraph{Gradient alignment.} Without a differentiable simulator, true gradients are not available. The closest reference is automatic differentiation through the global model, a reference of convenience whose own accuracy is not guaranteed. Even this reference covers only part of our objective: the value term operates in the local latent space and cannot be differentiated through the global model, so we compare the directions of the reward and dynamics terms. On the G1 world model, our decoupled gradient reaches a cosine similarity of 0.52 with this reference, against 0.35 for a non-decoupled Dreamer-style rollout and 0.08 for the local model before online fine-tuning: decoupling yields a $1.5\times$ better-aligned gradient.

\paragraph{Compute cost.} Backpropagating through the global model takes 45.6~GB and 1.5~s for a single state over a 16-step chunk, i.e., 128 H100 GPUs for our 128 actors. Our method takes 59~ms and 3.6~GB for the same work: $25\times$ faster on $128\times$ fewer GPUs for this gradient computation (the full G1 training uses 8 H100 per seed; Table~\ref{tab:hyperparameters}).

\clearpage

\subsection{Extended Quantitative Results: Per Task Sample and Time Efficiency} \label{apx:per_task_efficiency}
\begin{figure}[H]
    \centering
    \begin{subfigure}{0.9\linewidth}
        \centering
        \includegraphics[width=\linewidth]{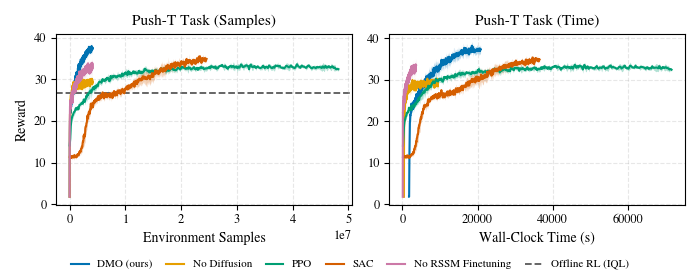}
        \caption{}
        \label{fig:efficiency_pusht_a}
    \end{subfigure}

    \begin{subfigure}{0.9\linewidth}
        \centering
        \includegraphics[width=\linewidth]{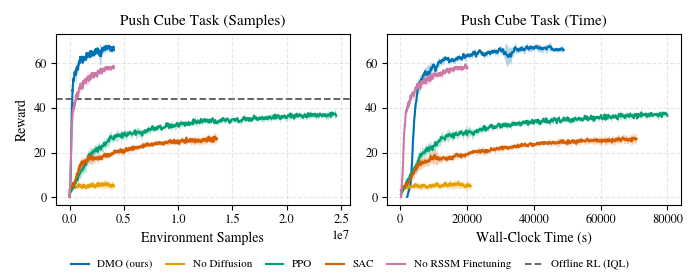}
        \caption{}
        \label{fig:efficiency_pusht_b}
    \end{subfigure}

    \begin{subfigure}{0.9\linewidth}
        \centering
        \includegraphics[width=\linewidth]{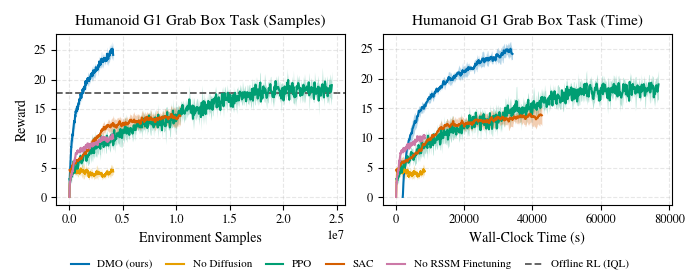}
        \caption{}
        \label{fig:efficiency_g1_c}
    \end{subfigure}

    \caption{(a) Efficiency comparison on the Push-T task: sample efficiency of DMO (4M samples) versus PPO (48M samples), SAC, and the No Diffusion and No RSSM Finetuning ablations (left), and corresponding time efficiency comparison (right). (b) Efficiency comparison on the Push Cube task: sample efficiency of DMO (4M samples) versus PPO (25M samples), SAC, and the No Diffusion and No RSSM Finetuning ablations (left), and corresponding time efficiency comparison (right). (c) Efficiency comparison on the G1 Grab Box task: sample efficiency of DMO (4M samples) versus PPO (25M samples), SAC, and the No Diffusion and No RSSM Finetuning ablations (left), and corresponding time efficiency comparison (right). The dashed line is the best evaluation of offline RL (IQL), which uses no environment samples. Every experiment was run with 4 seeds.}
    \label{fig:combined_efficiency_pusht_pushcube}
\end{figure}

All learning curves report the return of the training episodes: the undiscounted sum, over a 64-step episode, of the global reward model's per-frame predictions, as a running mean over recent episodes, with stochastic actions and without the entropy bonus. Each episode starts from a context drawn from the recorded data (a single frame on Push-T, a 32-frame window on Push Cube and G1). In Fig.~\ref{fig:aggregated_efficiency}, rewards are min-max normalized per task over all logged values; sample counts and wall-clock times are divided by the largest value reached in that task; curves are averaged over seeds and then over tasks; the shaded band is one standard deviation across tasks.

\subsection{The Algorithm}
Given the described global and local models, we present our complete procedure in Algorithm \ref{alg:main}. To ensure stable policy updates, we use the SAPO version of the DMO algorithm, which incorporates the Soft Analytic Policy Optimization (SAPO) objective \cite{xing2024stabilizing}. %, a variant of DMO that incorporates the Soft Analytic Policy Optimization (SAPO) objective \cite{xing2024stabilizing}. 
Specifically, we optimize the policy parameters $\boldsymbol{\theta}$ to maximize the entropy-regularized expected return:
%
% \begin{equation}
% \label{fo:5}
% \begin{split}
% \mathcal{L}^{\text{DMO-SAPO}}_\pi(\boldsymbol{\theta}) & := \mathbb{E}_{\tau \sim \pi_\theta, f}\Bigg[\sum_{h=1}^{H-1} \gamma^h 
%     \big(r\left(s_h, a_h\right) + \alpha \mathcal{H}_{\pi} \left[ a_h \mid s_h \right]\big) 
%       \\ &+ \gamma^H V^{\pi_\theta}_{\boldsymbol{\psi}}\left(s_H\right)\Bigg],
% \end{split}
% \end{equation}
\begin{equation}
\label{fo:5}
\begin{split}
\mathcal{L}^{\text{DMO-SAPO}}_\pi(\boldsymbol{\theta}) & := \mathbb{E}_{\tau \sim \pi_\theta, f^\rightarrow}\Bigg[\sum_{h=0}^{H-1} \gamma^h 
    \big(\hat{R}_\phi\left(l_{h+1}\right) + \alpha\, \tilde{\mathcal{H}}_h\big) 
      + \gamma^H V^{\pi_\theta}_{\boldsymbol{\psi}}\left(l_H\right)\Bigg],
\end{split}
\end{equation}
where $l_h$ is the local latent state at the $h$-th frame of the trajectory generated by the global model $f^\rightarrow$, $\hat{R}_\phi$ is the reward head of the local model, and $V^{\pi_\theta}_{\boldsymbol{\psi}}$ is the mean of an ensemble of 10 critics. As in SAPO~\cite{xing2024stabilizing}, the entropy term is normalized with the target entropy $\bar{\mathcal{H}}<0$ (Table~\ref{tab:hyperparameters}): $\tilde{\mathcal{H}}_h := (\hat{\mathcal{H}}_h + |\bar{\mathcal{H}}|)/(2|\bar{\mathcal{H}}|)$, where $\hat{\mathcal{H}}_h = -\log \pi_\theta(a_h \mid l_h)$ is a single-sample estimate of the entropy of the action distribution (a tanh-squashed, scaled Gaussian), and the temperature $\alpha$ is tuned automatically by minimizing $\alpha(\hat{\mathcal{H}}_h - \bar{\mathcal{H}})$ with the unnormalized estimate. If an episode ends at step $d$ inside the unroll, the sum in Eq.~\eqref{fo:5} stops there and ends with $\gamma^{d+1} V^{\pi_\theta}_{\boldsymbol{\psi}}(l_{d+1})$ instead of $\gamma^H V^{\pi_\theta}_{\boldsymbol{\psi}}(l_H)$, and a new sum starts.

The critic is an ensemble of 10 networks $V_{\boldsymbol{\psi}_k}$, trained by minimizing the temporal difference error against $\lambda$-returns:
\begin{align}
    \mathcal{L}_V(\boldsymbol{\psi}):=\sum_{k=1}^{10}\sum_{h=0}^{H-1}\left\|V_{\boldsymbol{\psi}_k}\left(l_h\right)-\hat{V}\left(l_h\right)\right\|_2^2, \label{fo:2}
\end{align}
where, for $t=0,\dots,H-1$:
\begin{align}
    \begin{split}
    V_h\left(l_t\right)&:=\sum_{n=t}^{t+h-1} \gamma^{n-t} \tilde{r}_n+\gamma^{h} V^{\min}_{\boldsymbol{\psi}}\left(l_{t+h}\right) \\
    \hat{V}\left(l_t\right)&:=(1-\lambda)\left[\sum_{h=1}^{H-t-1} \lambda^{h-1} V_h\left(l_t\right)\right]+\lambda^{H-t-1} V_{H-t}\left(l_t\right)
    \end{split}
 \end{align}
Here, $\tilde{r}_n := r_n + \alpha\tilde{\mathcal{H}}_n$ is the global reward model's prediction $r_n$ plus the normalized entropy bonus, and $V^{\min}_{\boldsymbol{\psi}}$ is the minimum of two ensemble members drawn at random at each step; no separate target network is used. If an episode ends at step $d<H-1$, $H$ is replaced by $d+1$ for $t\le d$.

When unrolling the DreamerV4 global world model, we employ KV Caching to significantly reduce inference time.
To initialize the context of the DreamerV4 world model during RL training, we do not start from a single frame. Instead, we sample a short history window of length $L_{\text{init}}=32$ from the real trajectory dataset and pre-fill the key-value (KV) caches of the Transformer. This provides the model with sufficient temporal context to resolve potential ambiguities (e.g., momentary occlusions of the object) before the policy begins its interaction.

\begin{algorithm}[t]
\caption{DMO-SAPO with global and local world models} \label{alg:main}

\begin{algorithmic} 
\State Offline training of the DIAMOND or DreamerV4 diffusion world model $f^\rightarrow$ and global reward model $r$
\State Offline pretraining of the local DreamerV3 world model $f^\leftarrow_\phi$ (including its reward head $\hat{R}_\phi$)
\State Fill the replay buffer $\mathcal{B}$ with transitions generated by the initial policy in $f^\rightarrow$
\For{epoch = 1 to N}
    \State \# Backward model learning
    \For{model mini epoch ($10{,}000$ updates in the first epoch, $64$ afterwards)}
        \State Sample sequences of transitions $(o, a, r)$ from $\mathcal{B}$
        \State$\boldsymbol{\phi} \leftarrow \boldsymbol{\phi}-\eta_{\boldsymbol{\phi}} \nabla_\phi\mathcal{L}_{f^\leftarrow}(\boldsymbol{\phi})$, where $\mathcal{L}_{f^\leftarrow}$ is the DreamerV3 model learning loss
    \EndFor
    \State \# Actor and critic learning
    \State $total\_reward \gets 0$
    \State $l_0 \gets$ latent state of $f^\leftarrow_\phi$ for the current frame $o_0$
    \For{$h = 0$ to $H-1$}
        \State $a_h \sim \pi_\theta(\cdot \mid l_h)$, \; $\hat{\mathcal{H}}_h \gets -\log\pi_\theta(a_h \mid l_h)$, \; $\tilde{\mathcal{H}}_h \gets (\hat{\mathcal{H}}_h + |\bar{\mathcal{H}}|)/(2|\bar{\mathcal{H}}|)$
        \State $o_{h+1} \gets f^\rightarrow(o_h, a_h)$ 
        (where $f^\rightarrow$ is the DIAMOND or the DreamerV4 diffusion global world model)
        \State $r_h \gets r(o_{h+1}, a_h)$ (where $r$ is the global reward model, only used to learn the local reward and the value function)
        \State Store $(o_h, a_h, r_h)$ in $\mathcal{B}$
        \State $l_{h+1} \gets$ latent state of $f^\leftarrow_\phi$ for $o_{h+1}$, from $\text{encoder}(o_{h+1})$ and its recurrent state
        \State $\hat{R}_h \gets \hat{R}_\phi(l_{h+1})$
        \State $total\_reward \gets total\_reward + \gamma^h\big(\hat{R}_h + \alpha\,\tilde{\mathcal{H}}_h\big)$
    \EndFor
    \State Compute $\mathcal{L}^{\text{DMO-SAPO}}_\pi(\boldsymbol{\theta})$ from $total\_reward$ and $\gamma^H V_{\boldsymbol{\psi}}^{\pi_\theta}(l_{H})$
    \State Compute $\mathcal{L}_V(\boldsymbol{\psi})$ from $(l_0, \dots, l_{H})$ and $(\tilde{r}_0, \dots, \tilde{r}_{H-1})$
    \State Use $\frac{\partial f^\leftarrow_\phi(l, a)}{\partial l} \bigg|_{(l_{h},a_{h})}$ and $\frac{\partial f^\leftarrow_\phi(l, a)}{\partial a} \bigg|_{(l_{h},a_{h})}$ to approximate $\frac{\partial f(o, a)}{\partial o} \bigg|_{(o_{h},a_{h})}$ and $\frac{\partial f(o, a)}{\partial a} \bigg|_{(o_{h},a_{h})}$, $h = 0, \dots, H-1$, during the following backward pass.
    \State $\nabla_\theta\mathcal{L}^{\text{DMO-SAPO}}_\pi(\boldsymbol{\theta}) \gets \text{backward}(\mathcal{L}^{\text{DMO-SAPO}}_\pi(\boldsymbol{\theta}))$
    
    \State $\boldsymbol{\theta} \leftarrow \boldsymbol{\theta}+\eta_{\boldsymbol{\theta}} \nabla_\theta\mathcal{L}^{\text{DMO-SAPO}}_\pi(\boldsymbol{\theta})$
    \State Update $\log\alpha$ by a gradient descent step on $\frac{1}{H}\sum_{h=0}^{H-1}\alpha\,(\hat{\mathcal{H}}_h - \bar{\mathcal{H}})$
    \State $\boldsymbol{\psi} \leftarrow \boldsymbol{\psi}-\eta_{\boldsymbol{\psi}} \nabla_\psi\mathcal{L}_V(\boldsymbol{\psi})$
\EndFor
\end{algorithmic}
\end{algorithm}

\clearpage

\subsection{Additional Real-World Deployments and Behaviors}
\label{apx:additional_deployments}
\begin{figure}[h]
    \centering
    \includegraphics[width=\linewidth]{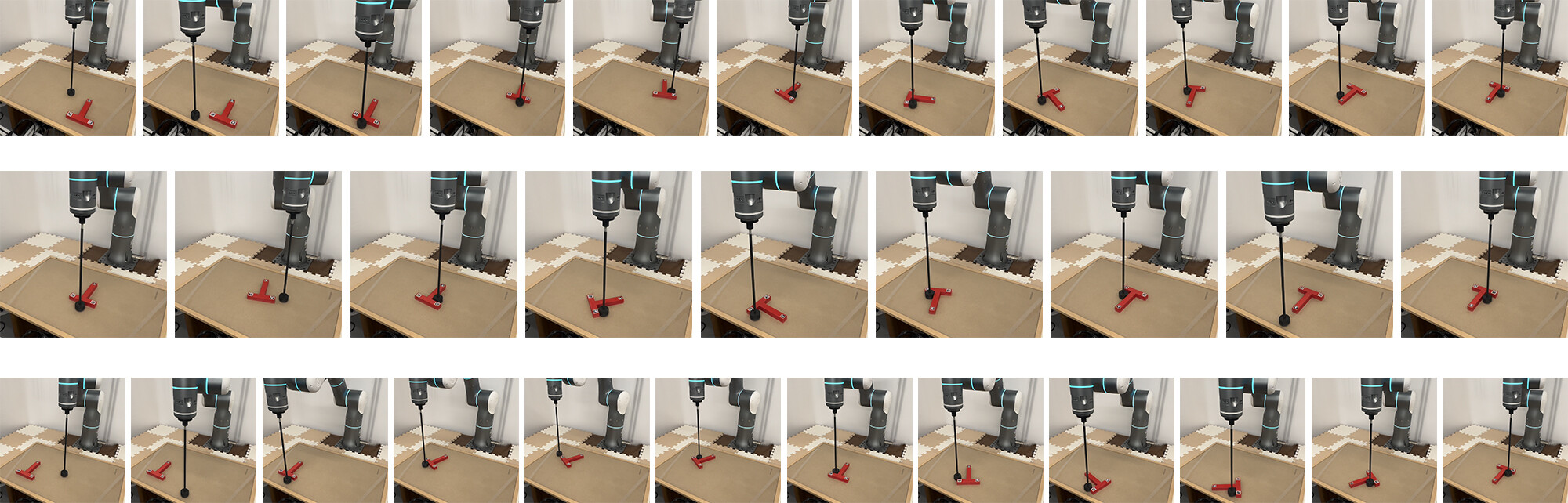}
    \caption{Three real-robot Push-T trajectories executed by the policy learned with our approach.}
    \label{fig:pusht-deployment}
\end{figure}

\begin{figure}[h]
    \centering
    \includegraphics[width=\linewidth]{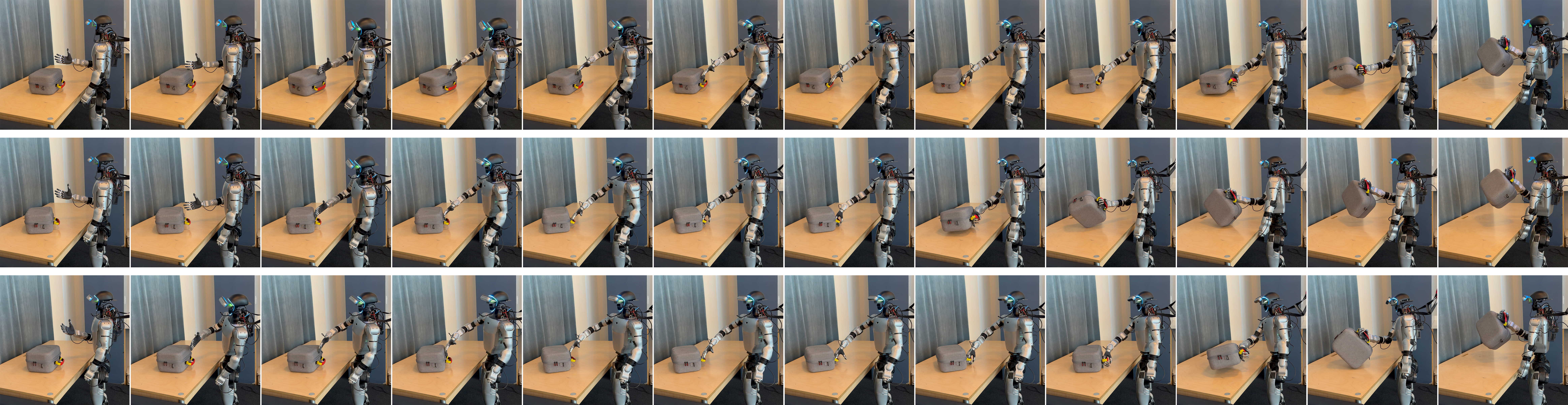}
    \caption{Three real-robot G1 Grab Box trajectories executed by the policy learned with our approach.}
    \label{fig:g1_grab_box_deployment}
\end{figure}

\begin{figure}[h]
    \centering
    \includegraphics[width=\linewidth]{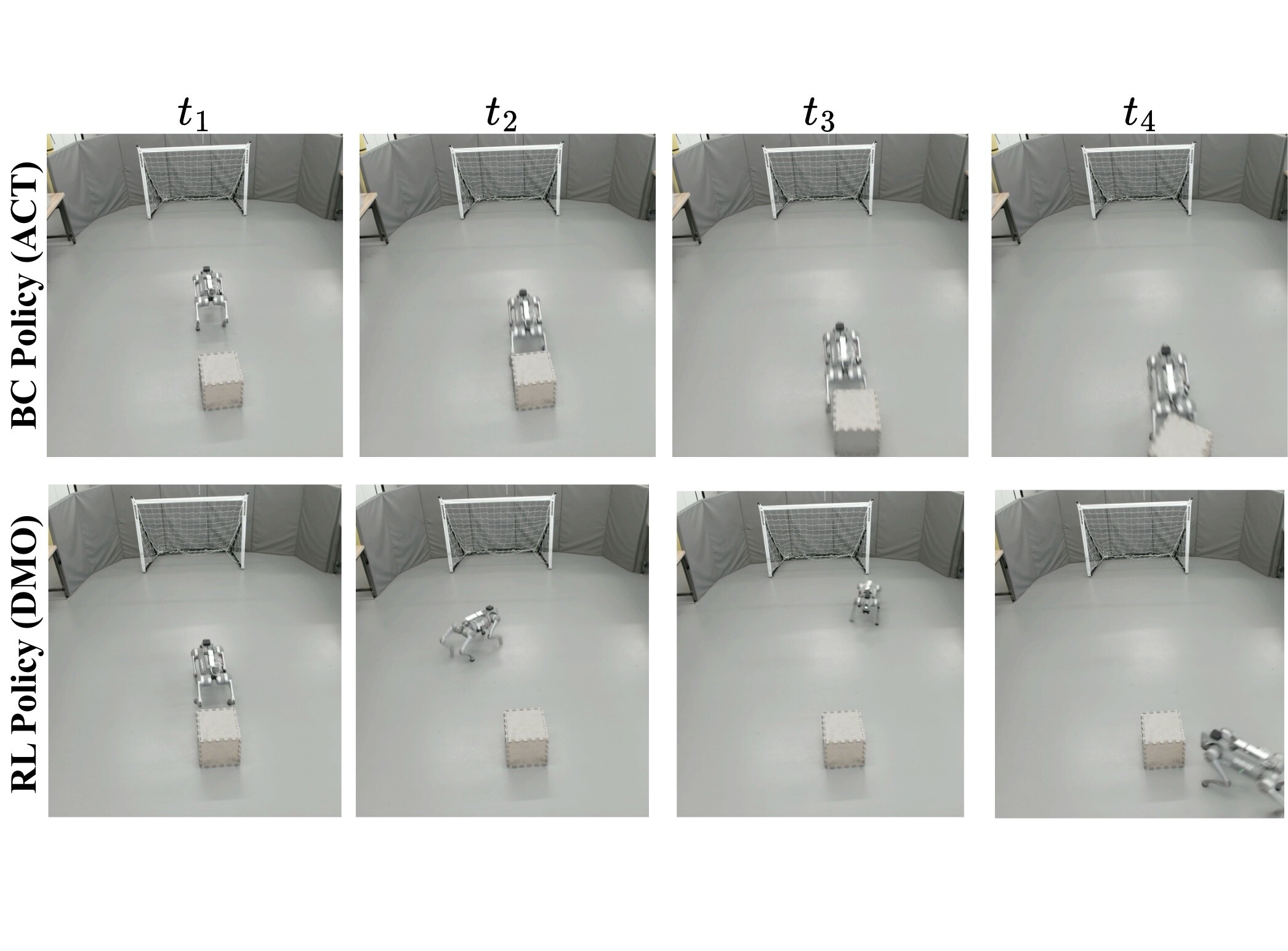}
    \caption{Comparison of search strategies under partial observability. \textbf{Top:} When the cube is out of view, the BC policy (ACT) executes a naive backward retreat, which fails to locate the object if it is directly behind. \textbf{Bottom:} The DMO policy exhibits an emergent active search behavior, rotating to visually scan the environment and successfully localize the target.}
    \label{fig:push_cube_sequence_search}
\end{figure}

As shown in Fig.~\ref{fig:push_cube_sequence_search}, DMO exhibits an emergent active search strategy when the object is out of view. Specifically, the agent retreats to the goal area and executes a full 360$^\circ$ yaw rotation to scan the environment. In contrast, the BC policy (first row of Fig.~\ref{fig:push_cube_sequence_search}) simply moves backward when the cube is lost. This blind retraction is brittle (if the cube is located behind the robot, the agent drags it farther away) and reflects the poor search strategy in the demonstration subset of the dataset. However, RL leverages random exploration during training to discover more effective behaviors.

\begin{figure}[h]
    \centering
    \includegraphics[width=\linewidth]{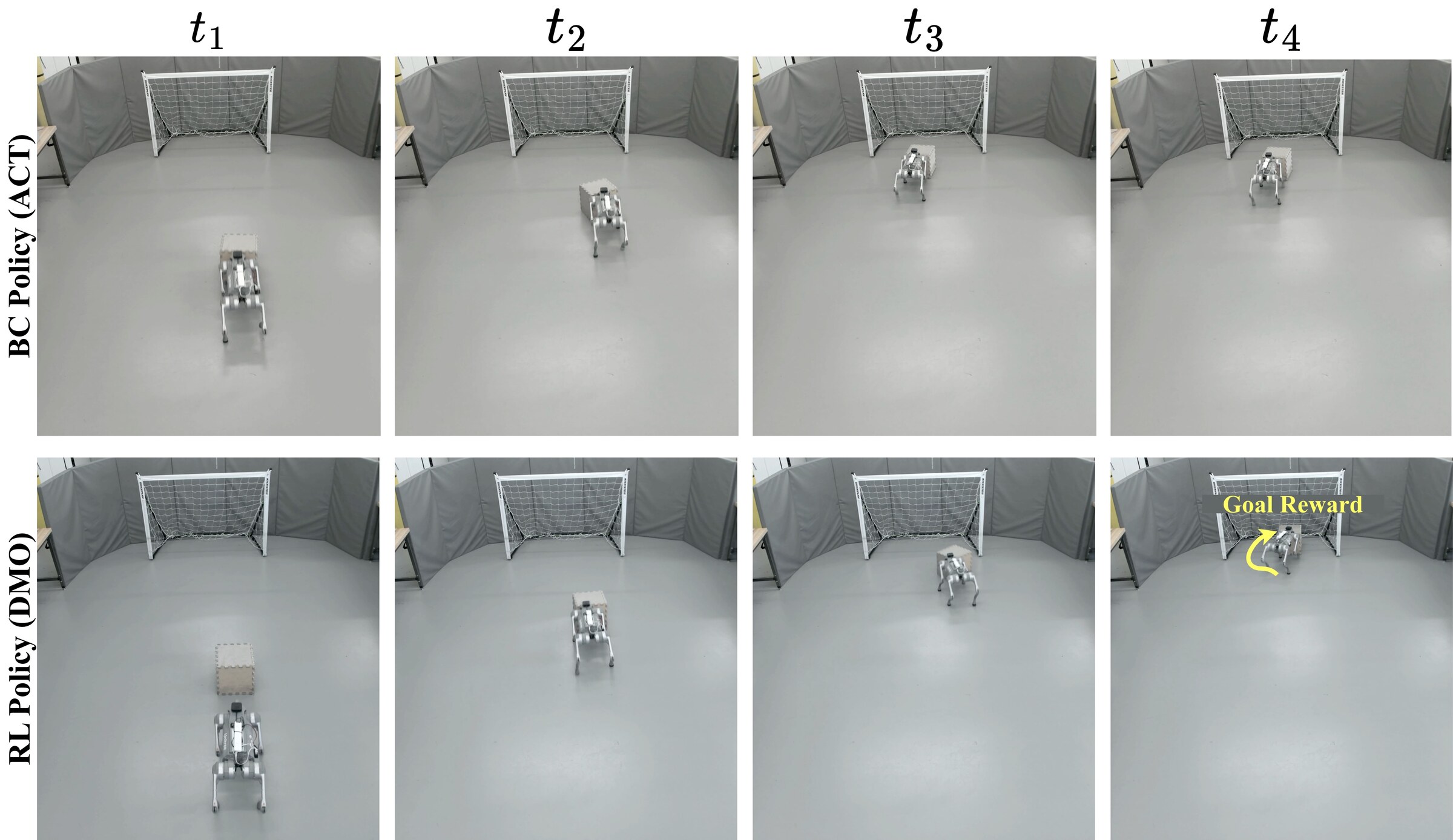}
    \caption{Task completion comparison. \textbf{Top:} The Behavior Cloning (ACT) policy successfully approaches the cube but stops pushing before the object enters the goal, reflecting the sub-optimal demonstration distribution. \textbf{Bottom:} Our RL Policy (DMO) generalizes beyond the demonstrations, learning to push the cube fully into the net to maximize the task reward.}
    \label{fig:push_cube_sequence_bc_goal}
\end{figure}

Then, as mentioned in Sec.~\ref{sec:results} and shown in Fig.~\ref{fig:push_cube_sequence_bc_goal}, the BC (ACT) policy consistently stops before fully pushing the cube into the goal, whereas DMO successfully completes the task by leveraging sparse successful examples with high terminal rewards.

\begin{figure}[h]
    \centering
    \includegraphics[width=\linewidth]{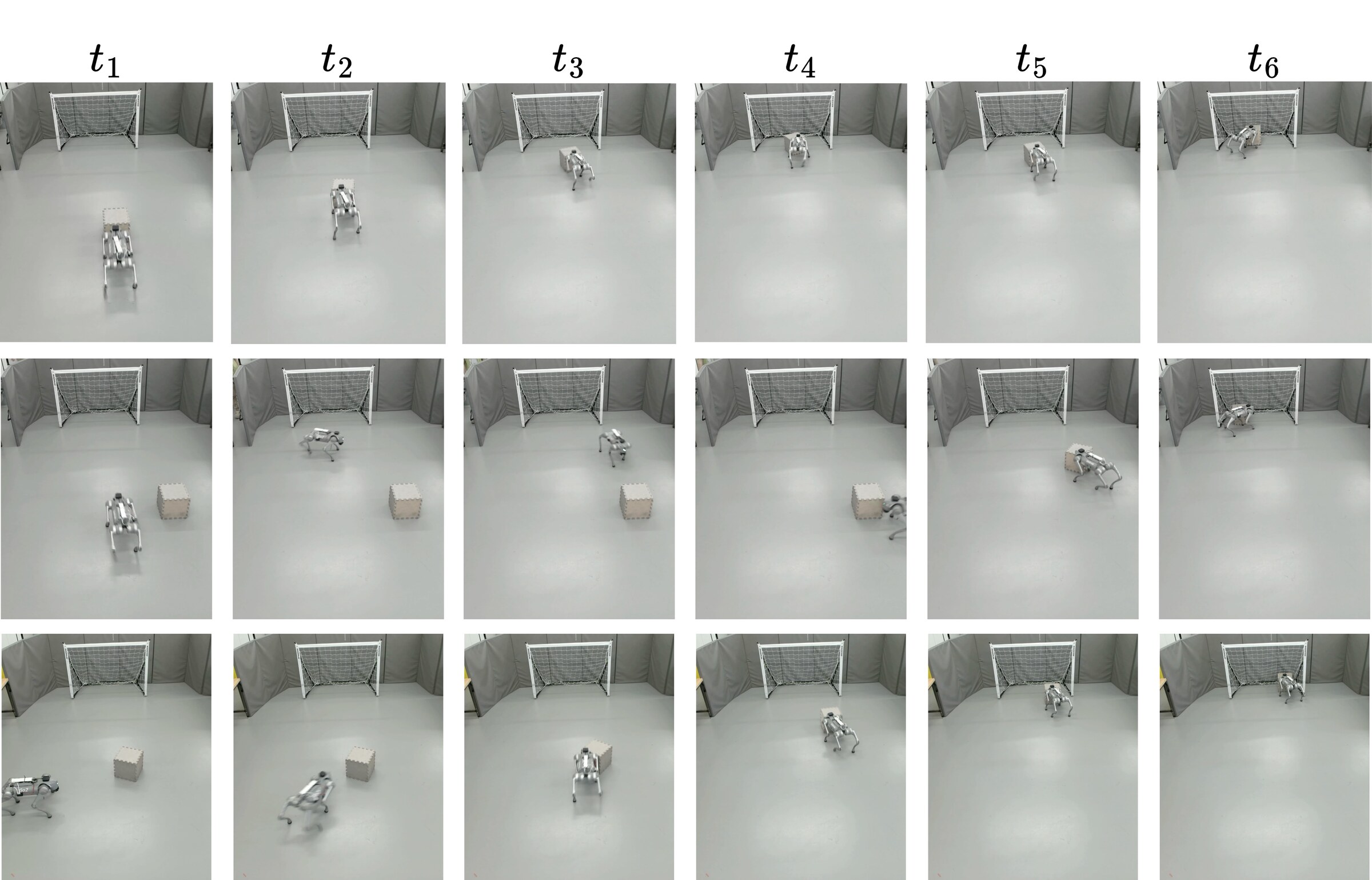}
    \caption{Real-world execution of the DMO-learned policy for the Go2 Push Cube task when using a different low-level policy than seen during training. \textbf{Top:} Standard successful push trajectory. \textbf{Middle:} Emergent active search behavior; when the object is initially out of view (or lost), the agent retreats and scans the environment to localize the cube before pushing. \textbf{Bottom:} Successful approach and push from a different initial configuration.}
    \label{fig:push_cube_sequence_dmo}
\end{figure}

Finally, we evaluated robustness to dynamics distribution shifts by replacing the low-level locomotion controller with a variant with a different response behavior and gait style (no Raibert heuristics in reward). As shown in Fig.~\ref{fig:push_cube_sequence_dmo}, without collecting new data or retraining the world models, the robot successfully pushes the cube into the goal. While PPO failed to generalize due to its higher entropy and less smooth action output, both DMO and BC maintained their performance. This result addresses a common critique of model-based methods—that models must be retrained for every specific robot configuration. Instead, this experiment demonstrates that the DMO framework correctly discovers stabilizing and smooth feedback strategies that are robust to variations in low-level dynamics. This finding implies that pretrained general-purpose world models may still produce useful policies for downstream tasks. A deeper investigation of this axis is the topic of our future research.

\clearpage

\subsection{Implementation Details and Hyperparameters}
\label{apx:hyperparameters}

\begin{table}[H]
    \centering
    \caption{\textbf{Hyperparameters.} Settings for PPO and DMO-SAPO across the three tasks (in the RL block, values that differ across tasks are given as Push-T / Push Cube / G1 Grab Box), along with the specific World Model architectures used in our framework. DIAMOND is used for Push-T, while DreamerV4 is used for both Push Cube and G1 Grab Box.}
    \label{tab:hyperparameters}
    \vspace{0.5em}
    
    % Left Column: RL Algorithms and Local World Model
    \begin{minipage}[t]{0.49\textwidth}
        \centering
        \resizebox{\linewidth}{!}{%
        \begin{tabular}{lcc}
        \toprule
        \multicolumn{3}{c}{\textbf{RL Algorithms (All Tasks)}} \\
        \midrule
        Parameter & PPO & DMO-SAPO \\
        \midrule
        Image Size (Input) & $64 \times 64$ & $64 \times 64$ \\
        Horizon ($H$) & 64 & 16 \\
        Gamma ($\gamma$) & 0.99 & 0.99 \\
        Lambda ($\lambda$) & 0.95 & 0.95 \\
        Mini-batch Size & 4096 & 1024 \\
        Mini-epochs & 4 & 8 \\
        Optimizer & Adam & AdamW \\
        Optimizer $(\beta_1, \beta_2)$ & (0.9, 0.999) & (0.7, 0.95) \\
        LR Schedule & Linear & Linear \\
        Learning Rate (Actor) & $5\mathrm{e}{-4}$ & $2\mathrm{e}{-3}$ \\
        Learning Rate (Critic) & $5\mathrm{e}{-4}$ & $5\mathrm{e}{-4}$ \\
        Learning Rate (Entropy) & -- & $5\mathrm{e}{-3}$ \\
        Entropy Coef / Target & 0.01 & $-6$ / $-6$ / $-58$ \\
        Grad Norm Clip & 0.5 & 1.0 \\
        Replay Buffer Size & -- & $10^6$ \\
        Num Actors & 256 & 128 \\
        Network & CNN + MLP & MLP [256, 128, 64] \\
        Activation & ReLU & SiLU \\
        Compute (per Seed) & 1 H100 / 4 H200 / 4 H200 & 1 H100 / 4 H200 / 8 H100 \\
        \midrule
        \multicolumn{3}{c}{\textbf{Local World Model (RSSM - DreamerV3 Style)}} \\
        \midrule
        Parameter & Push-T & Push Cube \& G1 Grab Box \\
        \midrule
        Learning Rate & $1\mathrm{e}{-4}$ & $1\mathrm{e}{-4}$ \\
        Recurrent State Size & 1024 & 4096 \\
        Stochastic/Discrete Size & 32 & 32 \\
        Dense Units & 640 & 1024 \\
        Channels Multiplier & 48 & 96 \\
        MLP Layers & 3 & 5 \\
        KL Scales (Dyn/Rep/Free) & 0.5 / 0.1 / 1.0 & 0.5 / 0.1 / 1.0 \\
        Reward Bins / Range & 255 / symlog $[-20, 20]$ & 255 / symlog $[-20, 20]$ \\
        \bottomrule
        \end{tabular}%
        }
    \end{minipage}\hfill
    % Right Column: Global World Models
    \begin{minipage}[t]{0.49\textwidth}
        \centering
        \resizebox{\linewidth}{!}{%
        \begin{tabular}{lcc}
        \toprule
        \multicolumn{3}{c}{\textbf{Global World Model (DreamerV4)}} \\
        \midrule
        Parameter & Push Cube & G1 Grab Box \\
        \midrule
        Image Size (Generation) & $128 \times 128$ & $256 \times 256$ \\
        Patch Size & 16 & 16 \\
        Enc/Dec Model Dim & 768 & 1024 \\
        Dynamics Model Dim & 1536 & 1536 \\
        Num Latents / Bottleneck & 256 / 16 & 256 / 32 \\
        Enc/Dec/Dyn Layers & 12 / 12 / 24 & 16 / 16 / 24 \\
        Enc/Dec/Dyn Heads & 12 / 12 / 24 & 16 / 16 / 24 \\
        Diffusion Steps & 4 & 4 \\
        Context Length (RL) & 96 & 32 \\
        Per Rank Batch Size (Tok. / Dyn.) & 2 / 1 & 1 / 1 \\
        Compute (Tok. / Dyn.) & 24 H200 / 64 H100 & 64 H100 / 64 H100 \\
        Gradient Accum. Steps (Tok. / Dyn.) & 5 / 4 & 4 / 4 \\
        Effective Batch Size (Tok. / Dyn.) & 240 / 256 & 256 / 256 \\
        Training Updates (Tok. / Dyn.) & 130k / 29k & 45k / 52k \\
        \midrule
        \multicolumn{3}{c}{\textbf{Global World Model (DIAMOND - Push-T Only)}} \\
        \midrule
        Parameter & \multicolumn{2}{c}{Value} \\
        \midrule
        Image Size (Generation) & \multicolumn{2}{c}{$64 \times 64$} \\
        Diffusion Steps & \multicolumn{2}{c}{3} \\
        Conditioning Steps & \multicolumn{2}{c}{4} \\
        UNet Channels & \multicolumn{2}{c}{[64, 64, 64, 64]} \\
        UNet Depths & \multicolumn{2}{c}{[2, 2, 2, 2]} \\
        Attn Depths & \multicolumn{2}{c}{[0, 0, 0, 0]} \\
        Cond Channels & \multicolumn{2}{c}{256} \\
        Sigma Distribution & \multicolumn{2}{c}{$\mu=-0.4, \sigma=1.2$} \\
        Optimizer & \multicolumn{2}{c}{AdamW ($\beta=0.999$)} \\
        Learning Rate & \multicolumn{2}{c}{$1\mathrm{e}{-4}$} \\
        Batch Size & \multicolumn{2}{c}{2 (Accum Grad 4 $=$ 8)} \\
        Compute & \multicolumn{2}{c}{1 RTX4090} \\
        \bottomrule
        \end{tabular}%
        }
    \end{minipage}
\end{table}

\paragraph{Behavior cloning baseline.} The BC baseline is the LeRobot~\cite{cadene2024lerobot} implementation of ACT~\cite{zhao2023learningfinegrainedbimanualmanipulation}, with a ResNet-18 backbone, hidden dimension 512 and a KL weight of 10. On G1, it uses 4 encoder and 1 decoder layers and a 32-dimensional latent, is trained on 231 demonstrations (25,960 frames at 10~Hz) for 200k steps at batch size 256 and learning rate $10^{-4}$, and executes chunks of 20 actions open-loop at 10~Hz. On Push Cube, it is trained on 1,200 episodes (37,971 frames at 5~Hz) of $64\times64$ images for 700k steps at batch size 128 and learning rate $10^{-5}$; it predicts 25 actions but executes only the first and re-plans at every step.

\subsection{CEM-MPC with the Global World Model}
\label{apx:mpc_go2}

We additionally implemented a Cross-Entropy Method (CEM) model-predictive controller for the Go2 Push Cube task using the same global world model and reward model employed during RL training. This experiment was designed to assess whether the learned global model could be used directly for online planning, without policy distillation.

In order to make planning as efficient as possible, we applied several aggressive optimizations. First, we reduced the number of diffusion steps from 4 to 1. Second, we used only a single outer-loop CEM refinement iteration for each action executed on the robot. Third, we exploited the DreamerV4 dynamics model in chunk-diffusion mode, enabled by the diffusion-forcing-style training procedure, which supports both autoregressive one-step diffusion and joint diffusion over multiple future steps.

Despite these optimizations, online planning remained prohibitively expensive for real-time control. In practice, achieving real-time inference still required at least 16 H200 GPUs for 128 parallel rollouts. These results, therefore, support our central motivation for policy learning: while large world models can provide accurate forward predictions, direct test-time planning with them remains too computationally demanding, and memorizing the resulting behavior into a policy, as enabled by DMO, remains necessary.

%===============================================================================

\end{document}